\documentclass[a4,journal]{IEEEtran}
\DeclareUnicodeCharacter{2212}{-}
\usepackage{wrapfig}
\usepackage{amsmath,amsthm,amssymb}
\usepackage{array}
\usepackage{graphicx}
\usepackage{makecell}
\usepackage{tabularx}

\usepackage{caption}

\usepackage{hyperref}
\usepackage{comment}
\usepackage[letterpaper,top=2cm,bottom=2cm,left=2cm,right=2cm,marginparwidth=1.75cm]{geometry}
\usepackage{hyperref}

\title{SAFE: a SAR Feature Extractor based on self-supervised learning and masked Siamese ViTs}
\title{SAFE: a SAR Feature Extractor based on self-supervised learning and masked Siamese ViTs}
\author{\IEEEauthorblockN{Max Muzeau$^{1,2}$}, \IEEEauthorblockN{Joana Frontera-Pons$^{1}$},
    \IEEEauthorblockN{Chengfang Ren$^{2}$},  
   \IEEEauthorblockN{Jean-Philippe Ovarlez$^{1,2}$},
   \thanks{$^{1}$ DEMR, ONERA, Universit\'e Paris-Saclay, F-91120 Palaiseau, France}
     \thanks{$^{2}$ SONDRA, CentraleSup\'elec, Universit\'e Paris-Saclay, F-91192 Gif-sur-Yvette, France.}
    
}
\date{}
\begin{document}

\maketitle

\begin{abstract}
    Due to its all-weather and day-and-night capabilities, Synthetic Aperture Radar imagery is essential for various applications such as disaster management, earth monitoring, change detection and target recognition. However, the scarcity of labeled SAR data limits the performance of most deep learning algorithms. To address this issue, we propose a novel self-supervised learning framework based on masked Siamese Vision Transformers to create a General SAR Feature Extractor coined SAFE. Our method leverages contrastive learning principles to train a model on unlabeled SAR data, extracting robust and generalizable features. SAFE is applicable across multiple SAR acquisition modes and resolutions. We introduce tailored data augmentation techniques specific to SAR imagery, such as sub-aperture decomposition and despeckling. Comprehensive evaluations on various downstream tasks, including few-shot classification, segmentation, visualization, and pattern detection, demonstrate the effectiveness and versatility of the proposed approach. Our network competes with or surpasses other state-of-the-art methods in few-shot classification and segmentation tasks, even without being trained on the sensors used for the evaluation.
\end{abstract}
\markboth{Submitted to IEEE Transactions on Geoscience and Remote Sensing}{}

\section{Introduction}
Synthetic Aperture Radar (SAR) imagery has proven advantageous across various applications, such as environmental monitoring \cite{komarov2020assimilation}, disaster management \cite{tay2020rapid}, military surveillance \cite{kechagias2021automatic} and urban planning \cite{rambour2020interferometric}. Unlike optical imaging, SAR can operate day and night and is weather-independent. It can see through clouds and, in some cases, penetrate vegetation and soil. This ability to consistently acquire reliable data makes SAR a vital tool for applications that require uninterrupted monitoring. Furthermore, SAR provides high spatial resolution independent of the distance between the sensor and the target \cite{moreira2013tutorial}. The transmitted wavelengths vary roughly from 1cm to 1m, depending on the desired applications. As the wavelengths increase, the resolution decreases, but there is more surface penetration. Short wavelengths such as 3cm (X-band) can be used to detect vehicles \cite{8124934}, whereas larger wavelengths such as 30cm (L-band and P-band) are used more generally to monitor biomass and vegetation \cite{chang2022application}. During emission and reception, the wavelength can be polarized horizontally or vertically, leading to a total of 4 polarized channels $[HH, HV, VH, VV]$\footnote{HV means a horizontal polarization during the emission and a vertical one during the reception} called Polarimetric SAR (PolSAR) data. Polarization diversity will highlight specific types of structures \cite{lee2017polarimetric}.

Although the proliferation of satellites equipped with SAR sensors has exponentially increased the volume of SAR data available, the scarcity of labeled images remains a significant obstacle. Labeled data is crucial for training supervised learning algorithms predominantly used in deep learning models. While all these algorithms rely on labeled data to learn meaningful patterns, annotating them is often labor-intensive, expensive, and impractical for large-scale datasets.

Self-Supervised Learning (SSL) offers a promising solution to exploit the vast amount of unlabeled SAR data by enabling models to learn valuable representations. SSL algorithms are typically implemented through pretext tasks that do not require manual labeling, such as predicting the rotation of an image or identifying whether two augmented views of the same image are similar \cite{chen2020simple}. It has been used in SAR images for multiple applications such as despeckling \cite{dalsasso2021if,dalsasso2022self}, anomaly detection \cite{max_aae}, super-resolution \cite{muzeauimpact}, and target recognition \cite{pei2023self}. Among various SSL approaches, contrastive learning has emerged as particularly effective. Contrastive learning involves training a model to distinguish between similar and dissimilar data pairs in a latent space, encouraging the model to learn robust and discriminative features. This technique employs data augmentation, transforming each data to create multiple views, generating positive or negative pairs. These augmented pairs are then fed into an encoder that maps them into a latent space. The model uses a similarity measure, such as cosine similarity or Euclidean distance, to evaluate the closeness of these representations. 

Given the potential of SSL, our objective is to develop a general SAR feature extractor that can be applied to a wide range of tasks. Although such general feature extractors have been successfully implemented in other modalities, such as standard images and natural language processing, their application to SAR imagery remains largely unexplored. Our goal is to create a network capable of extracting pertinent features from unseen SAR data, thereby enhancing the versatility and effectiveness of SAR-based applications. 

This paper proposes a novel SSL framework using contrastive learning principles to build a robust and general SAR Feature Extractor (SAFE). The proposed approach aims to fill the current literature gap for future SAR SSL model advancements. Our contributions are threefold: 
\begin{itemize}
    \item  To introduce a novel SSL framework tailored for SAR imagery
    \item To demonstrate the effectiveness of masked Siamese ViTs for SAR feature extraction on real SAR data
    \item To provide comprehensive evaluations on various downstream tasks, showcasing the versatility and robustness of our proposed method
\end{itemize}
The paper is organized as follows. In Section \ref{sec:sota}, we introduce the notion of SSL and provide a summary of the methods employed for feature extraction, along with their applications in SAR imaging. In Section \ref{sec:method}, we describe our proposed architecture, along with the different data augmentation methods. In Section \ref{sec:experiments}, we present multiple experiments applying SAFE to segmentation, few-shot classification, feature visualization, and pattern detection. Conclusions and perspectives are discussed in Section \ref{sec:conclusions}. The Python code created to make the network and the different experiments, along with the weights, are available at the URL: \href{https://github.com/muzmax/SAFE}{https://github.com/muzmax/SAFE}.

\section{Related works}
\label{sec:sota}
The related work is decomposed into three parts. First, there is a short explanation of the SSL principle. Then, there is an overview of the self-supervised feature extractor algorithm. Finally, some adaptations of these methods for SAR images are explained.

\subsection{Self-supervised learning principle}
\label{sec:concepts}
\begin{figure}
   \centering
   \includegraphics[width=\linewidth]{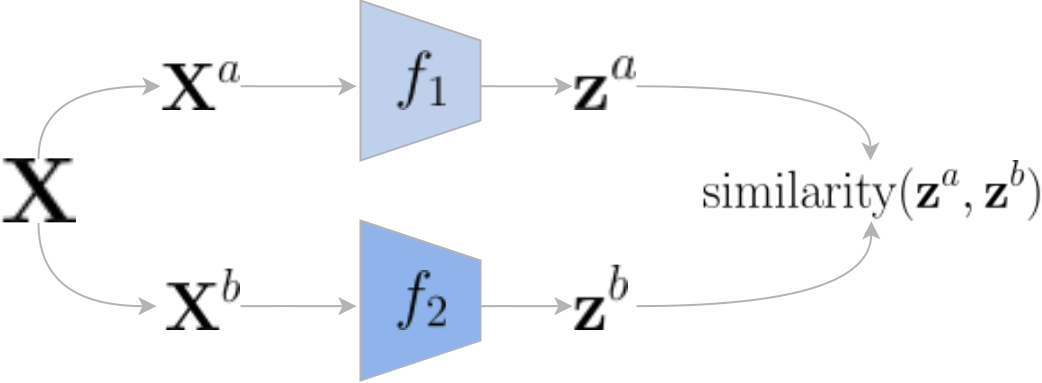}
  \caption{Principle of SSL with Siamese networks. Notations are the one explained in \ref{sec:concepts}}
  \label{fig:siamese}
\end{figure}

Self-supervision aims to teach a model to create a useful representation of data without relying on labels \cite{bengio2013representation}. The key idea is to design pretext tasks, such as predicting a rotation angle, to train the network. Doing so will make it understand the inherent structure, relationships, or semantics \cite{balestriero2023cookbook}. The most common self-supervised method in computer vision is based on a type of contrastive learning method called Siamese networks \cite{bromley1993signature}, which consists of the following principle: Two encoders $f_1$ and $f_2$ are used to compress images in a latent space. From an image $\mathbf{X}$, two augmented views $\mathbf{X}^a$ and $\mathbf{X}^b$ are generated and fed to the networks. Then, the latent vectors $\mathbf{z}^a = f_1(\mathbf{X}^a)$ and $\mathbf{z}^b = f_2(\mathbf{X}^b)$ are checked for similarity or dissimilarity (see Fig.\ref{fig:siamese})\cite{he2020momentum,chen2020simple,caron2021emerging,caron2020unsupervised}. The performance of Siamese networks depends on two main factors. Data augmentation methods will be the most important information provided to the network to decide whether two images have the same latent features, so they must be chosen carefully to fit the desired result perfectly. The other factor is the training loss and regularization methods to ensure the latent space does not collapse.

\subsection{Self-supervised feature extraction}
The earliest algorithms that showed the potential to close the gap between supervised and SSL are only a few years old. One of the first and simplest is SimCLR \cite{chen2020simple}. For a given image, two augmented versions are created using augmentations such as crop, colour distortion, and blur. Then, for a given batch, a ResNet encodes every image in a latent vector to find each positive pair (the rest of the images are implicitly defined as negative pairs in the loss function).
Another architecture called MoCo \cite{he2020momentum} uses a more memory-efficient strategy to compute negative samples by storing them in a memory bank. The other difference is using a momentum encoder, a moving-averaged version of the main encoder.\\

Due to the challenges associated with defining negative pairs without supervision and the computational complexity involved, recent neural networks only use positive pairs. This change is possible using new loss functions and training methods, such as the momentum encoder mentioned earlier. This is the case of SwAV \cite{caron2020unsupervised}, and this algorithm also uses training strategies that increase performance compared to previous approaches. One method uses a multi-view consistency objective, where the model is trained to bring different augmentations of the same image closer together in the learned representation space. The second method is a mechanism to learn self-supervised cluster assignments. The model is trained to predict the cluster assignments of one view from the cluster assignments of another view.\\

The last developed algorithms \cite{caron2021emerging,assran2022masked} rely on Vision Transformers (ViTS) \cite{dosovitskiy2020image}. This architecture allows for capturing long-range dependencies from the first attention layer \cite{vaswani2017attention}, as opposed to convolutional networks, where the receptive field gradually increases. Because ViTs process images in a patch-wise manner, it adds interesting properties. For instance, in the Masked Siamese Networks (MSN) \cite{assran2022masked} architecture, a portion of the encoded patch is randomly masked as an augmentation for the training phase. Another property of ViTs is the adaptability to variable-sized inputs. They were initially developed to make neural language processing \cite{vaswani2017attention}, so the architecture had to be able to take sentences of different shapes as inputs. It can be useful to take images of different shapes in input, especially when dealing with SAR data.\\
\subsection{SSL in SAR}
\label{subsec:SSL_SAR}
Most of these algorithms have been used extensively in multispectral and hyperspectral earth observation \cite{wang2022self}, but the applications are more limited regarding SAR imagery. For polarimetric synthetic aperture radar, the paper \cite{ren2021mutual} trains a convolutional network with contrastive learning. Different representation features are extracted from a PolSAR image, each consisting of an augmented view. The pretext task of the neural network will be to force the same encoding, no matter the representation. Used features are the coherency matrix, the Pauli decomposition, the $H, A, \alpha$ decomposition and the Freeman decomposition \cite{lopez2021basic}. Another method focuses on Graph Convolutional Networks (GCN) \cite{kipf2016semi} for mono-channel SAR classification \cite{liu2022contrastive}. A standard ResNet is trained in a self-supervised manner using BYOL \cite{grill2020bootstrap} to extract relevant features. Jointly, a classification algorithm \cite{wu2020multiscale} is trained to extract multiscale and multidirectional features. Both extracted features will be processed conjointly by a GCN and a fully connected network for an optimized classification. In a more standard fashion, the paper \cite{park2022homography} trains the network MoCo \cite{he2020momentum} with a homography transformation for data augmentation \cite{detone2016deep}. It is chosen as an easily applicable augmentation suitable for SAR instead of most optical data augmentation methods.\\

Most of the Self-supervised methods, particularly contrastive learning algorithms, employ the MSTAR dataset as it is one of the few labeled datasets available. The paper \cite{xu2021adversarial} uses a MoCo-like algorithm to create a network that is robust to external perturbation of the different types of vehicles. The supervised defences \cite{zhang2019theoretically,wong2020fast} are combined with MoCo to achieve this result. In the MSTAR dataset, vehicles are sorted by angles. The article \cite{wen2021rotation} leverages this information to obtain rotation awareness. A two-branch convolutional neural network has to predict the rotational pattern between two images of the same class in addition to the label. Another paper \cite{pei2023self} simply trains a SimCLR-like network \cite{chen2020simple} on preprocessed MSTAR data. Jointly, each trained network is evaluated on a linear classification task to see which network will give the optimal performance for the desired task. Sometimes, neural networks can classify targets based solely on the background or on the shadow of the image. To counter this, the paper \cite{peng2023learning} randomly changes the clutter of an image while preserving the target. A convolution network encodes two images of the same target with a different clutter. In addition to the standard cross-entropy loss, a channel-weighted mean square error is added, forcing the model to align the features to learn invariant target representation. Because the MSTAR dataset might not be optimal for training a neural network, researchers developed the SAMPLE dataset \cite{lewis2019sar}, which consists of a subset of the MSTAR dataset and their synthetically generated counterparts. The generated images have different noise levels, the clutter differs, and the target aspect can significantly vary. The paper \cite{geng2023target} showed that a ResNet that performs well on the MSTAR dataset could see its accuracy drop by more than 30\% when the clutter background is modified in the evaluation phase only. This is why, like the previous article, a clutter transfer augmentation is implemented. Because most optical data augmentations are not suited for SAR imagery, they use a physic-based SAR augmentation first proposed in \cite{agarwal2023sparse} and apply it to the SAMPLE dataset in addition to the clutter transfer. Based on the header information of each image and on the assumption that the data acquisition is in spotlight mode \cite{moreira2013tutorial}, a view from a slightly different azimuth angle is made via interpolation in the phase-history domain.\\

\begin{figure}
   \centering
   \includegraphics[width=\linewidth]{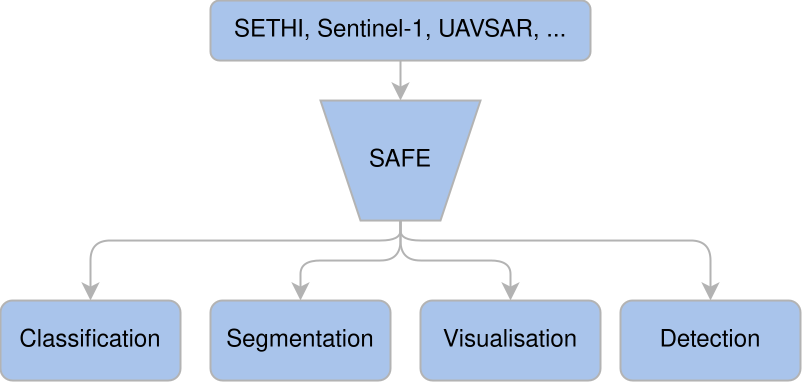} 
  \caption{Principle of our SAR Feature Extractor (SAFE). One network can extract meaningful features from different SAR acquisition methods. These features can then be used to do many downstream tasks.}
  \label{fig:safe}
\end{figure}
\begin{figure*}[t]
    \centering
    \includegraphics[width=\linewidth]{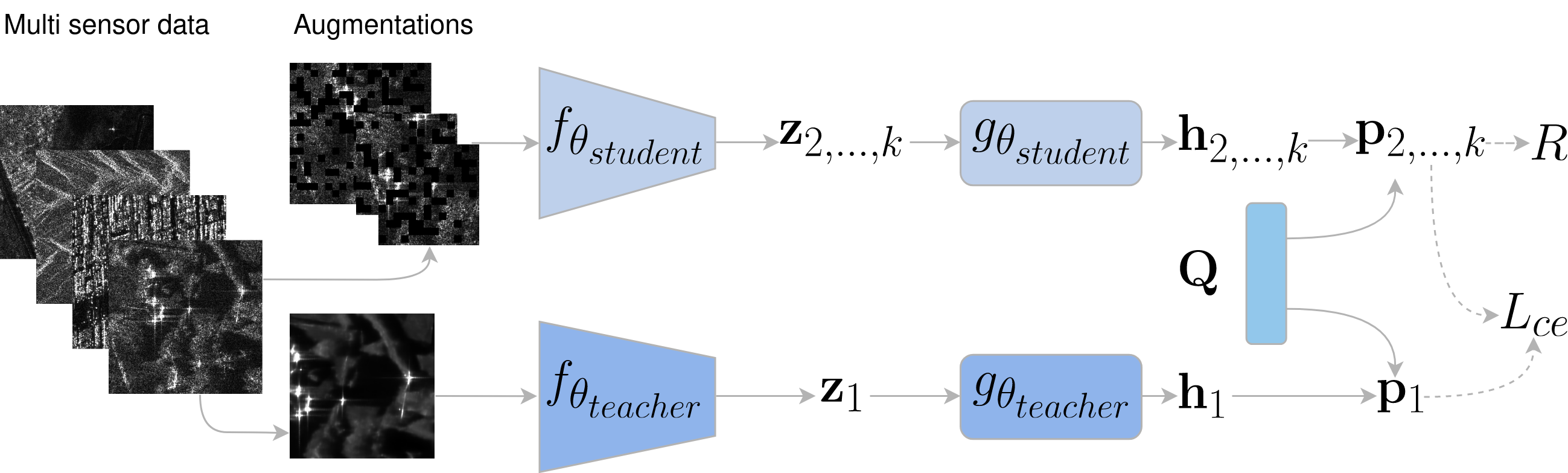} 
    \caption{Model architecture for the training phase. The notations are detailed in \ref{subsec:model}.}
    \label{fig:archi}
\end{figure*}
\section{Proposed Method}
\label{sec:method}
In contrast with the methods presented in \ref{subsec:SSL_SAR}, our objective is to develop a general SAR feature extractor. This extractor is designed for any acquisition mode and resolution, as well as both mono-channel SAR and PolSAR amplitude data. As illustrated in Fig.~\ref{fig:safe}, our ambition is to train a SAR feature extractor on various SAR data types and then use it as a backbone for diverse applications. First, the architecture of the feature extractor is presented along with its training methodology. Then, each data augmentation method is explained. 
\subsection{Model}
\label{subsec:model}
\textbf{Architecture.} Our method displayed in Fig.~\ref{fig:archi} is similar to the principle explained in \ref{sec:concepts}. The methodology employs two distinct networks, denoted as $f_{\theta_{student}}$ and $f_{\theta_{teacher}}$, functioning as feature extractors.  Given an image $\mathbf{X}\in \mathbb{R}^{h\times w\times c}$, these networks generate a feature vector $\mathbf{z} = f_{\theta}(\mathbf{X})\in \mathbb{R}^{d_f}$. With $h,w,c$ and $d_f$ being the height, width, channel number and latent space dimension. While having the same ViT architecture \cite{dosovitskiy2020image}, these networks have different weights. The applied augmentations require the use of a transformer-based framework, which is suitable for varied image dimensions. This allows input images of different sizes to be taken in without resizing them to the same shape, thus retaining the image's intrinsic characteristics. In a similar manner to ViTs, two Multilayer Perceptron (MLP) heads, $g_{\theta_{student}}$ and $g_{\theta_{teacher}}$, project the extracted features to a vector $\mathbf{h} = g(\mathbf{z})\in \mathbb{R}^{d_g}$, $d_g$ being the projected space dimension. Discarding the MLP head after the training phase helps create a robust feature extractor, mitigating the risk of over-fitting to specific tasks.   
 Subsequently, these vectors are projected and normalized against a prototype set $\mathbf{Q} = [\mathbf{q}_1,\mathbf{q}_2,\ldots,\mathbf{q}_n]\in \mathbb{R}^{d_g \times n}$, yielding a cosine similarity measure $\mathbf{s} = [s_1,s_2,\ldots,s_n]^T \in \mathbb{R}^{n}$ as per the equation:
\begin{equation}
 s_i = \frac{\mathbf{q}_i^T \mathbf{h}}{\left\| \mathbf{q}_i\right\|_2 \, \left\| \mathbf{h}\right\|_2}, \, \forall i\in [|1;n|].
\end{equation}
Relying on the similarity vector $\mathbf{s}$ instead of $\mathbf{h}$ allows for better clustering capacities. It also makes the feature extractor more efficient for few-shots downstream tasks \cite{assran2022masked,wang2020generalizing}. Finally, a softmax is computed such that:
\begin{equation}
\label{eq:softmax}
\mathbf{p} = \mathrm{softmax}\left(\displaystyle\frac{\mathbf{s}}{\tau}\right)\, ,
\end{equation}
where $\tau$ is a temperature to soften or sharpen the distribution. With the intent of aligning $f_{\theta{student}}$'s representations more closely with the one of $f_{\theta{teacher}}$, we set $\tau$ smaller for the teacher than for the student. By doing so, it forces it to have a sharper prediction. 
 \\
\textbf{Training.} The network is trained on patches of SAR images with either one or full polarizations. For each patch $\mathbf{X}$, $k$ augmented views $\mathbf{X}_{1,\ldots,k}$ are generated. The first image $\mathbf{X}_1$ is encoded via the teacher network, whereas the student network encodes the subsequent images. For a batch of size $b$ we note $\mathbf{p}^l_{i,j}$ the l-component of the $n$-vector computed in \eqref{eq:softmax} for $j$-th augmented view of the $i$-th image of a batch.

\begin{figure*}[t]
    \centering
    \includegraphics[width=0.93\linewidth]{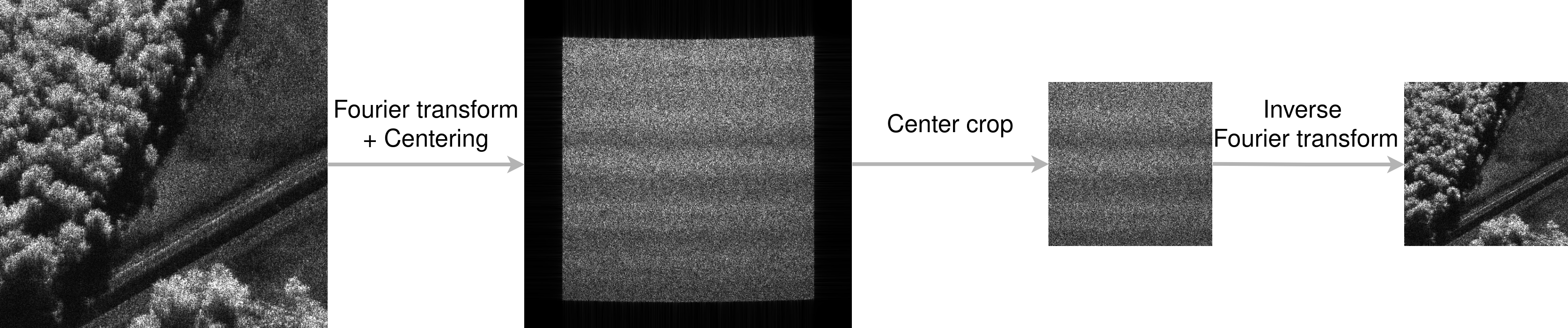}  
    \caption{Method for the sub-aperture augmentation of SAR images}
    \label{fig:subsample}
\end{figure*}

The training incorporates two distinct loss functions. The first is a cross-entropy loss, defined as:
\begin{equation}
L_{ce} = \frac{1}{b~(k-1)} \sum \limits_{i = 1}^b \sum \limits_{j = 2}^k \sum_{l=1}^n  -\mathbf{p}_{i,1}^l~\log{\mathbf{p}_{i,j}^l}\, ,
\end{equation}
This formulation ensures that all vectors $\mathbf{p}_{2,...,k}$ encoded within a batch are aligned with $\mathbf{p}_1$, the vector encoded by the teacher network. Following, a mean-entropy maximization regularizer loss is employed \cite{assran2021semi}, forcing the network to utilize every prototype in $\mathbf{Q}$ throughout a batch:
\begin{equation}
R = -\sum_{l=1}^{n}\overline{\mathbf{p}}^l ~\log~ \overline{\mathbf{p}}^l  \text{ with } \overline{\mathbf{p}} = \frac{1}{b~(k-1)} \sum \limits_{i = 1}^b \sum \limits_{j = 2}^k \mathbf{p}_{i,j} \, ,  
\end{equation}
This approach makes the network robust to imbalanced datasets \cite{balestriero2023cookbook} and prevents it from collapsing. The final loss is computed such that $L =L_{ce} + \lambda R$. The parameter $\lambda$ is a positive number used to weight the importance of $R$. The networks $f_{\theta_{teacher}}$ and $g_{\theta_{teacher}}$ are not updated during the backpropagation, a moving average is used instead :
\begin{equation}
\theta_{teacher} \leftarrow m\, \theta_{teacher} + (1-m)\,  \theta_{student}.
\end{equation}
As with the previous regularization, it prevents the network from collapsing.

\subsection{Data augmentation}
\label{subsec:augmentations}
A crucial part of the method involves identifying relevant data augmentation techniques. The mapping of different images into similar or dissimilar feature vectors will be based on the augmentations. Given that most self-supervised models are created for optical images, their augmentations often do not translate well to SAR images. SAR images have a constant resolution regardless of the distance between the sensor and the scene. For instance, using a random Gaussian blur would force a representation that is independent of many resolutions, but in our case, the diversity in resolution is limited by the number of sensors. Two augmentations that account for most network performances are color jittering and masking parts of the image. Either through cropping or token suppression at the initial layer of a ViT \cite{chen2020simple,assran2022masked}. The color jittering, based on RGB normalized values, is incompatible with SAR images. Whereas the cropping and token suppression are independent of the dynamic range or structure of the image and can thus be directly applied to SAR. The selected augmentations, tailored for SAR images regardless of their polarization number, are the following:\\

\textbf{Normalization.}
Entities equipped with SAR acquisition systems use algorithms to calibrate the intensity of collected data. Despite these calibrations, differences in dynamic range can still occur. This may be due to target clipping during quantization to obtain a better dynamic range, as well as variations in sensor response to the same structures. The resolution of the sensor affects the number of backscatterers combined in a single pixel cell, and the wavelength of the sensor also impacts the intensity of the backscattering signal. These factors can lead to differences in intensity normalization across different sensors, resolutions, and frequency bands. To address these variations and ensure similar intensity, we preprocess the data to normalize the dynamic range across various sensors using the following equation:
\begin{equation}
\tilde{\mathbf{X}} = (\log \mathbf{X} - m_s) / (M_s - m_s) \, ,
\end{equation}
where $\mathbf{X}$ and $\tilde{\mathbf{X}}$ denote the amplitude image and the normalized image respectively. The log-amplitude of $\mathbf{X}$ is clipped between $\left[ m_s; M_s\right]$ where $m_s$ and $M_s$ are respectively the minimum and the maximum normalization parameters specific to the log-scale SAR system $s$. Hence, the matrix $\tilde{\mathbf{X}}$ takes values between 0 and 1. The logarithmic operation is employed to compress the dynamic range, addressing the high-intensity values of bright backscatterers compared to lower-intensity scatterers.\\

\textbf{Log-amplitude shift.}
After normalization, we introduce a log-amplitude shift similar to the colour jittering often used in optic imagery such that $\mathbf{x}_{shifted} = \mathbf{x}+B$, where $B\sim\mathcal{U}(a,b)$ is uniformly distributed between $a$ and $b$ . Unlike optical imagery, where intensity may be influenced by factors such as incident illumination, atmospheric conditions, and surface properties, SAR intensity primarily reflects the electromagnetic properties of the observed objects. This parameter should not be modified too importantly. With this slight bias, we encourage the network to prioritize structural understanding, which is something that is usually not taken into account with algorithms that rely solely on a statistical model.\\

\textbf{Global and local crops.}
We employ random global and local cropping from an image patch to generate similar samples. Doing so forces the network to learn a similarity measure based on the spatial proximity of the data. Having many local crops also boosts performance for a low additional computational cost.\\

\textbf{Token masking.}
In the ViT, the initial layer performs non-overlapping convolutions, extracting high-level features for subsequent attention layers. The number of convolution feature maps $d_f$ determines the token size, producing $n$ vectors $[\mathbf{v}_1,\mathbf{v}_2,...,\mathbf{v}_n]\in \mathbb{R}^{d_f\times n}$, with $n$ that depends on the size of the input image and on the kernel size. A percentage $p$ of the tokens are then masked to obtain the vectors $[\mathbf{v}^{'}_1,\mathbf{v}^{'}_2,...,\mathbf{v}^{'}_m]\in \mathbb{R}^{d_f\times m}$ with $m = pn$. Because the position information \cite{dosovitskiy2020image} is added before the masking, the spatial relation between tokens is preserved. By applying token masking, only a fraction of the encoded tokens are processed during training, enhancing memory efficiency and scalability with large datasets.\\

\textbf{Sub-aperture decomposition.}
Unlike optical images, SAR allows for subband and sub-aperture extraction, producing reduced-resolution images, as explained in \cite{JPO2003, Brekke2013}. SLC SAR data are complex-valued images with specific properties. Notably, we can compute the range and azimuth spectrum by applying the 2D Fourier transform. The original SAR spectrum should be centered around zero-Doppler and zero-azimuth frequencies, as shown in Fig. \ref{fig:subsample}. Then, the spectral image is low-pass filtered to obtain the desired resolution without zero-padding. This augmentation aims to learn invariance to resolution up to a certain scale. This allows, for example, to simulate a resolution change that we can observe between X and L band images.\\

\begin{figure}
\begin{minipage}[]{0.49\linewidth}
   \centering
   \includegraphics[width=\linewidth]{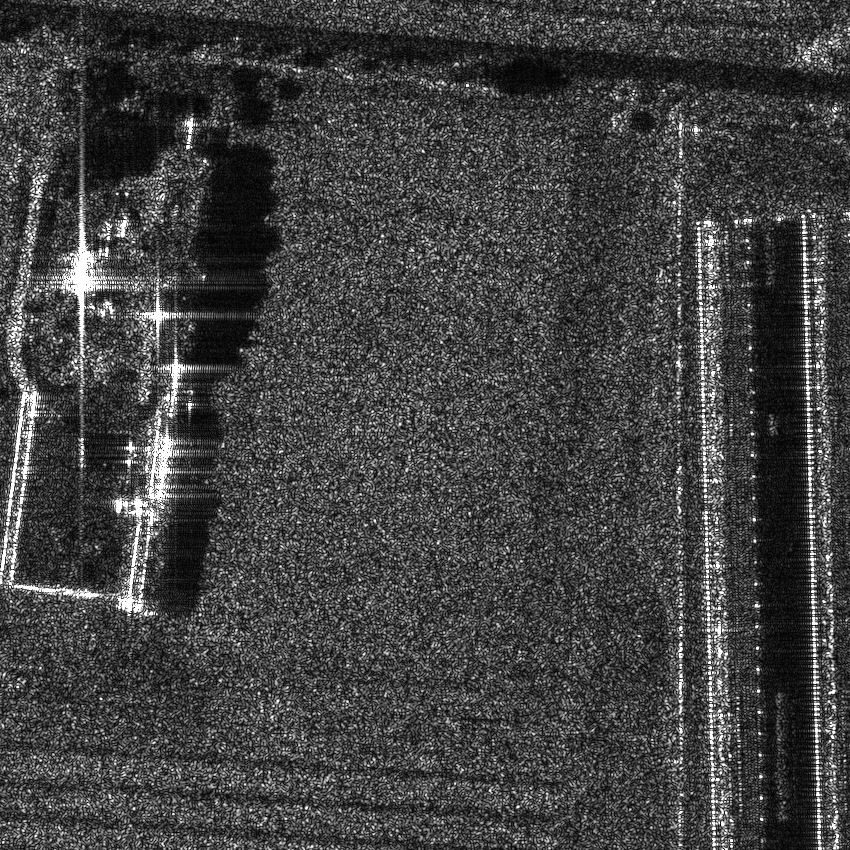}  
  \end{minipage}
    \hfill
  \begin{minipage}[]{0.49\linewidth}
   \centering
   \includegraphics[width=\linewidth]{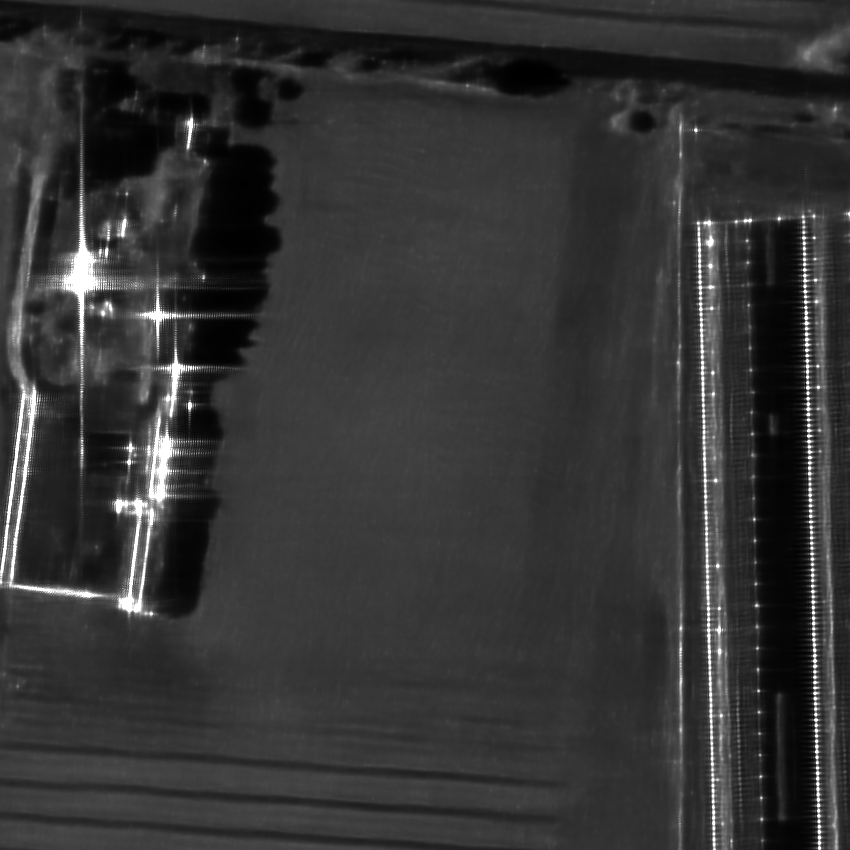}     
  \end{minipage}
  \caption{SLC image (left) and denoised SAR image (right) with MERLIN \cite{dalsasso2021if}}
  \label{fig:speckle}
\end{figure}

\textbf{Despeckling.}
SAR images are corrupted by a strong perturbation called speckle \cite{goodman1976some}. In addition to the known statistical distribution, spatial correlations are too complex to be considered by statistical filters \cite{deledalle2017mulog}. Traditional despeckling techniques needed data subsampling to whiten this noise, compromising image resolution. However, advances in Deep Learning have introduced algorithms capable of considering spatial correlation without resolution degradation \cite{dalsasso2022self}. In contrast to optical imagery, where noise is synthetically added as an augmentation, our approach leverages a noise-removal technique. Specifically, the MERLIN algorithm \cite{dalsasso2021if} is a U-Net model trained to reduce speckle without requiring ground truth, making it easy to train on any data (see Fig. \ref{fig:speckle}). We aim to make the network robust to speckle by simultaneously using despeckled and amplitude SLC images. We intentionally avoid a universal preprocessing despeckling step. Our model can inherently manage speckle interference after the training phase.\\

\textbf{Teacher and student augmentations.} The augmentation strategies for the teacher and student models differ. The teacher model processes images that are simpler to analyze, serving as a reference or `soft label` for the student model to emulate. Following the normalization process described previously, the teacher network uses a global crop of despeckled images. Conversely, the student augmentations consist of a combination of all the above augmentations, excluding the despeckling. This differentiated approach ensures that, while the teacher provides a clear and noise-free output, the student model learns from a broader variety of data representations, enhancing its adaptability and robustness.

\section{Experiments}
\label{sec:experiments}
To evaluate the performance of the feature extraction, we conduct tests on multiple downstream tasks, including segmentation on the AIR-PolSAR-Seg \cite{wang2022air} dataset, few-shot classification on the MSTAR dataset, visualization on the SETHI, UAV, and Sentinel-1 datasets, as well as pattern detection. Notably, the data used for segmentation, classification, and pattern recognition are not part of SAFE's training dataset. 
\subsection{Training dataset}
A diverse dataset encompassing images from various SAR acquisition modalities was used to train the network. This dataset includes:
\begin{itemize}
  \item A single image from the ONERA SETHI \cite{9078973} system operating in L-band, with $15000\times975$ pixels. This image primarily captures vegetation and has a resolution of 1m in both azimuth and range.
  \item Two images from the ONERA SETHI system in X-band, with sizes of $4800\times30000$ and $37400\times9230$ pixels. The first image is predominantly vegetated, whereas the second encompasses a blend of vegetation and urban structures. Each image has a resolution of 20cm in both range and azimuth.
  \item Three images from the Sentinel-1 \cite{sentinel1} sensor in C-band, with sizes of $7000\times10400$, $9000\times13000$, and $7200\times10000$ pixels. These images depict diversified landscapes: the first captures an urban area alongside a shore, the second portrays mountainous terrain, and the third features a combination of urban areas, vegetation, and agricultural plots. Each image has a resolution of 5m in both azimuth and range.
  \item An image from the \href{https://uavsar.jpl.nasa.gov/}{UAVSAR} sensor in L-band, with dimensions of $21700\times9786$ pixels. This image showcases a mix of urban areas and a shore. The resolution is 1.6m in range and 0.6m in azimuth.
\end{itemize}
Including images from different SAR bands (L, X, and C) and resolutions illustrates the comprehensive capability of the feature extractor to adapt and perform across a wide spectrum of SAR imaging conditions. This dataset has a variety of sensing modalities, but it also includes geographical landscapes, ranging from dense urban areas to vegetative regions and mountainous terrains, thereby ensuring that the trained network can generalize well across different SAR acquisition systems and environmental contexts. Additionally, the ONERA SETHI  and UAVSAR systems feature quad-polarizations $[HH, HV, VH, VV]$, while Sentinel-1 captures dual polarizations $[HV, VV]$ only. This polarization diversity impacts the training process; when using single-channel data, each polarization is fed into the network separately, increasing the dataset size by four or two. In contrast, for quad-polarization training, Sentinel-1 data is omitted, and the polarizations are concatenated into tensors of size $(H,W,4)$, with $H$ and $W$ indicating image dimensions and 4 representing the polarization channels.
\subsection{Implementation details}
Given the typically lower resolution of SAR images compared to their optical counterparts, a less complex network suffices, leading to the adjustment of certain parameters to smaller values compared to \cite{caron2021emerging}. Accordingly, the training images mentioned previously are segmented into non-overlapping patches of $100\times100$ pixels.

We employ the AdamW optimizer \cite{loshchilov2017decoupled} with a batch size of 512, warming up the learning rate linearly from 0 to $10^{−3}$ over the first 10 epochs, then applying a cosine decay schedule. Augmentations are applied as described in \ref{subsec:augmentations}, producing three local views of $32\times32$ pixels and two global views $64\times64$ pixels per image. The sub-aperture decomposition downsizes images from $100\times100$ to  $32\times32$ pixels. The token masking percentage is fixed at 0.3. The shift parameter follows a distribution $\mathcal{B}(0,0.3)$. Student and teacher networks use temperatures of $0.1$ and $0.04$, respectively. The teacher's network updates via an exponential moving average, starting with a momentum of $0.9995$, linearly increasing to $1$. Weight decay begins at $0.04$, rising to $0.4$ using a cosine schedule. The mean entropy maximization regularization weight $\lambda$ is set to 1. We utilize MLP projection heads with a $192$ output dimension and $256$ prototypes of the same dimension. The networks are trained for 600 epochs
\begin{figure*}[t]
    \centering
    \includegraphics[width=\linewidth]{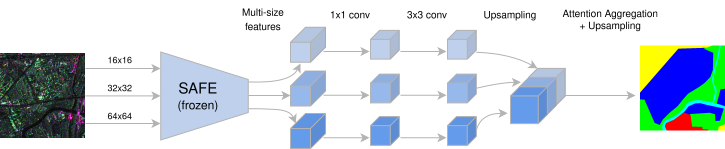} 
    \caption{Segmentation head integrated with SAFE. Initially, the image is partitioned into patches of sizes $16\times16$, $32\times32$, and $64\times64$ pixels, which are subsequently processed by the feature extractor. Each feature is processed by the same architectural framework but with distinct weights tailored to the specific patch size encoding. The extracted features are then merged using an attention layer to obtain the segmentation map.}
    \label{fig:seg}
\end{figure*}
\subsection{Segmentation}
\label{sec:seg}
In this part, we explore the use of a segmentation task to evaluate the quality of the extracted features, as well as the network's adaptability across various data types and ground surfaces for the four-channel feature extractor. Due to the nature of the Vit's output, which compresses an image into a vector and thus loses the spatial dimension, transformers are typically not used for segmentation tasks. Here, it is done not to achieve state-of-the-art segmentation results but rather as a means to assess the performance of the feature extraction process. Consequently, the architecture is designed such that the segmentation relies solely on the output of our feature extractor. Specifically, the segmentation will not perform effectively if the ViT fails to differentiate one surface type from another. A summary of the architecture is illustrated in Fig. \ref{fig:seg}\\

The evaluation utilizes the AIR-PolSAR-Seg dataset \cite{wang2022air}, which comprises $512 \times 512$ pixel images captured by a C-band spaceborne sensor with an $8 \times 8m$ resolution. Notably, this resolution is at least five times lower than the resolutions previously used for training the network. This dataset contains 500 fully polarized images $[HH, HV, VH, VV]$. To preserve spatial information using a ViT, each input image is segmented into patches of size $P \times P$, using a stride of $S$. These patches are then processed by the ViT to reconstruct a feature map analogous to the ones obtained through a convolutional network. Patch sizes $P\in\{16,32,64\}$ and a constant stride $S = 32$ were employed for our experiments. This consistent stride, coupled with padding, ensures that all feature maps maintain uniform spatial dimensions $(H/S, W/S)$, where $H$ and $W$ represent the height and width of the input image, respectively. Altering the patch dimensions enhances the diversity of the processed information, exploiting the ViT's capacity to handle varying input sizes effectively. \\

After feature map extraction with the 4 channels SAFE ViT, a convolutional architecture and an attention aggregation module inspired by the semantic segmentation network proposed in \cite{li2022a2} is applied. For each of the three distinct feature maps derived from varying patch sizes, the architecture employs a $1 \times 1$ kernel convolution to reduce dimensionality, followed by a $3 \times 3$ convolutional layer to refine the features. This is succeeded by three layers of upsampling convolutions. The processed feature maps are then integrated using the attention aggregation module, further processed by a $3 \times 3$ convolution, and ultimately resized to the original dimensions through bilinear upsampling. This segmentation head is trained over 100 epochs utilizing the AdamW optimizer and a cross-entropy loss. The initial learning rate was set at $3.125~10^{-5}$, which was reduced by a factor of 10 at epochs 40 and 80. The dataset was partitioned such that $75\%$ was allocated for training, and the remaining $25\%$ was used for validation. \\

To evaluate the segmentation, the following metrics are used: Overall Accuracy (OA), Average Accuracy (AA), Kappa Coefficient ($\mathrm{Kappa} $), and Mean Intersection over Union (mIoU). With:
\begin{equation}
    OA = \displaystyle\frac{\displaystyle\sum_{i=1}^{n} x_{ii}}{\displaystyle\sum_{i=1}^{n} \displaystyle\sum_{j=1}^{n} x_{ij}}\, ,
\end{equation}
where $x_{ii}$ are the correctly predicted pixels for class $i$, and $x_{ij}$ represents the total pixels predicted for class $i$ as class $j$,
\begin{equation}
    AA = \displaystyle\frac{1}{n} \sum_{i=1}^{n} \frac{x_{ii}}{\displaystyle\sum_{j=1}^{n} x_{ij}} \, ,
\end{equation}
where $n$ is the number of classes, and
\begin{equation}
    \mathrm{Kappa} = \frac{OA - p_e}{1 - p_e}\, ,
\end{equation}
with
\begin{equation}
    p_e = \displaystyle\sum_{i=1}^{n} \left(\frac{\displaystyle\sum_{j=1}^{n} x_{ij} \cdot \sum_{j=1}^{n} x_{ji}}{\left(\displaystyle\sum_{i=1}^{n} \sum_{j=1}^{n} x_{ij}\right)^2}\right)\, ,
\end{equation}
where $p_e$ is the expected agreement by chance.
\begin{equation}
    mIoU = \displaystyle\frac{1}{n} \displaystyle\sum_{i=1}^{n} \frac{x_{ii}}{\displaystyle\sum_{j=1}^{n} (x_{ij} + x_{ji} - x_{ii})}\, .
\end{equation}

Our experimental evaluation benchmarked the proposed `SAFE+seg' method against other recent deep learning architectures specifically developed for semantic segmentation. As indicated in Table~\ref{table:iou}, surfaces well-represented in the SAFE training dataset were effectively segmented. Our network achieved superior Intersection over Union (IoU) scores for `both industrial' and `natural' surfaces. However, although the results were satisfactory for `water' surfaces, they did not surpass those of other networks. Due to their low reflectivity, this performance discrepancy is attributed to the inherent simplicity of segmenting water surfaces in intensity SAR images. In the ViT processing stage, intensity information is mixed with others, complicating the segmentation task. A notable limitation of using SAFE for segmentation is its difficulty in clustering more abstract categories such as `Land Use' and `Other', resulting in lower IoU scores for these classes, adversely affecting the overall metrics. Despite this, as shown in Table~\ref{table:segmetrics}, our approach achieved the second-best OA, $0.24\%$ below the ANN and obtained the highest Kappa coefficient, surpassing the next best by $0.53\%$. Nevertheless, our method falls behind for the AA and mIoU, which are metrics that average scores across classes independently of pixel number, primarily due to its inability to segment the `Other' class effectively. These results underscore the adaptability of our feature extractor to unfamiliar data acquisition modalities, highlighting the necessity for training on surface types analogous to those encountered during testing. Future work could explore integrating the strengths of SAFE with established semantic segmentation networks to optimize performance.
\begin{table}[]
    \begin{center}
        \begin{minipage}{0.5\textwidth}
            \centering
            \resizebox{\textwidth}{!}
            {
            \begin{tabular}{c|c c c c c c} 
            
            \multicolumn{7}{c}{}\\
            \hline
            \\[-2ex] 
            Method  & Industrial & Natural & Land Use & Water & Housing & Other \\ [1ex] 
            \hline \\[-2ex] 
            DeepLav V3+ \cite{chen2018encoder} & 40.62 & 70.67 & 0.55 & \textbf{72.93} & 69.96 & 34.53 \\ 
                             
            ANN \cite{zhu2019asymmetric} & 41.23 & 72.92 & 0.97 & 75.95 & 68.40 & 56.01 \\
                            
            PSPNet \cite{zhao2017pyramid} & 33.99 & 72.31 & 0.93 & 76.51 & 68.07 & \textbf{57.07} \\
                            
            PSANet \cite{zhao2018psanet} & 40.70 & 69.46 & \textbf{1.33} & 69.46 & 68.75 & 32.68 \\
            SAFE+seg (ours) & \textbf{44.07} & \textbf{74.17} & 0.00 & 63.09 & \textbf{70.16} & 11.70 \\[1ex] 
            \hline
            \end{tabular}%
            }
        \end{minipage}          
    \end{center}
    \caption{IoU for Multi Category Segmentation}
    \label{table:iou}
    \end{table}
    
\begin{table}[]
    \begin{center}
        \begin{minipage}{0.43\textwidth}
            \centering
            \resizebox{\textwidth}{!}
            {
            \begin{tabular}{c|c c c c c c} 
            
            \multicolumn{5}{c}{}\\
            \hline
            \\[-2ex] 
            Method  & mIoU & OA & AA & Kappa\\ [1ex] 
            \hline \\[-2ex] 
            DeepLav V3+ \cite{chen2018encoder} & 48.21 & 76.81 & 63.55 & 64.92 \\ 
                             
            ANN \cite{zhu2019asymmetric} & \textbf{52.58} & \textbf{77.46} & \textbf{64.97} & 65.73 \\
                            
            PSPNet \cite{zhao2017pyramid} & 51.48 & 76.55 & 63.81 & 63.88 \\
                            
            PSANet \cite{zhao2018psanet} & 47.14 & 76.21 & 62.92 & 63.95 \\
            SAFE+seg (ours) & 43.87 & 77.22 & 54.43 & \textbf{66.26} \\[1ex] 
            \hline
            \end{tabular}%
            }
        \end{minipage}          
    \end{center}
    \caption{Metrics for Multi-Category Segmentation}
    \label{table:segmetrics}
    \end{table}

\subsection{MSTAR classification}
\begin{figure}[t]
    \centering
    \includegraphics[width=\linewidth]{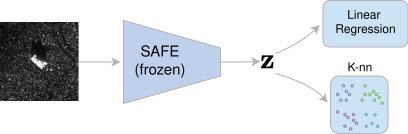}  
    \caption{Proposed Classification Method: the input image is passed through the feature extractor to obtain the feature vector $\mathbf{z}$. This feature vector is then used for classification by applying either a linear regression model or a k-NN algorithm. The feature extractor was not trained on the evaluation images in our experiments.}
    \label{fig:classif}
\end{figure}
Classification experiments were conducted using the one-channel feature extractor, with a particular focus on few-shot classification. The first set of experiments was performed on the MSTAR dataset, which comprises 30 cm X-band spotlight images of seven different types of vehicles and a calibration class. This data differs from the data on which the network was trained. Although some X-band data were included in the training set, they were acquired using stripmap mode with a different resolution. The specific military vehicles to be classified were not present in the training dataset, making this experiment a suitable test for demonstrating the adaptability of our network. Table~\ref{table:res} lists the different classes and the number of images per class. \\

\begin{table}[t]
\begin{center}
\begin{tabular}{|c |c| c| c| c|} 
 \multicolumn{5}{c}{MSTAR dataset} \\ [0.5ex] 
 \hline 
 Class & 2S1 & BRDM\_2 & BTR\_60 & D7 \\ [0.5ex] 
 \hline
 Number & 1664 & 1282 & 451 & 573 \\ [1ex] 
 \hline
 Class & T62 & ZIL131 & ZSU\_23\_4 & SLICY \\ [0.5ex] 
 \hline
 Number & 572 & 573 & 1401 & 2539 \\ [1ex] 
 \hline
\end{tabular}
\caption{\label{table:res}Image number for each class. The calibration class is called ``SLICY"; the others are vehicles.}
\end{center}
\end{table}

\begin{table*}[t]
    \centering
    \begin{tabularx}{\textwidth}{c | *{7}{>{\centering\arraybackslash}X}} 
    \multicolumn{8}{c}{}\\
    \hline
    \\[-1ex] 
    & \multicolumn{7}{c}{Number of labels per class} \\ [1ex] 
     Model &  1 &  2 &  3 &  4 &  5 &  10 &  50 \\ [1ex] 
    \hline \\[-2ex] 
    {\scriptsize SAFE+linear} &  {\scriptsize $\mathbf{69.64 ~ \pm ~ 0.39}$} &  {\scriptsize $\mathbf{78.06 ~ \pm ~ 0.13}$} &  {\scriptsize $\mathbf{89.64 ~ \pm ~ 0.14}$} &  {\scriptsize $\mathbf{90.13 ~ \pm ~ 0.09}$} &  {\scriptsize $\mathbf{90.80 ~ \pm ~ 0.08}$} &  {\scriptsize $\mathbf{97.31 ~ \pm ~ 0.03}$} &  {\scriptsize $\mathbf{98.56 ~ \pm ~ 0.07}$}\\[1ex] 
    
    {\scriptsize SAFE+k-NN} &  {\scriptsize $69.01$} &  {\scriptsize $76.73$} &  {\scriptsize $80.51$} &  {\scriptsize $88.80$} &  {\scriptsize $88.44$} &  {\scriptsize $94.59$} &  {\scriptsize $98.55$}\\[1ex] 

    {\scriptsize UMAP+k-NN\cite{mcinnes2018umap}} &  {\scriptsize $53.43 ~ \pm ~ 0.90$} &  {\scriptsize $52.34 ~ \pm ~ 0.85$} &  {\scriptsize $61.94 ~ \pm ~ 0.75$} &  {\scriptsize $62.56 ~ \pm ~ 0.61$} &  {\scriptsize $65.79 ~ \pm ~ 0.94$} &  {\scriptsize $69.27 ~ \pm ~ 0.79$} &  {\scriptsize $85.59 ~ \pm ~ 0.40$}\\[1ex] 

    {\scriptsize PCA+k-NN} &  {\scriptsize $48.08 ~ \pm ~0.24$} &  {\scriptsize $51.84 ~ \pm ~ 0.12$} &  {\scriptsize $62.67 ~ \pm ~ 0.16$} &  {\scriptsize $61.75 ~ \pm ~ 0.16$} &  {\scriptsize $63.68 ~ \pm ~ 0.15$} &  {\scriptsize $71.85 ~ \pm ~ 0.31$} &  {\scriptsize $91.04 ~ \pm ~ 0.17$}\\[1ex] 

    {\scriptsize ResNet18+lblsm\cite{inkawhich2021bridging}} &  {\scriptsize $36.67 ~ \pm ~ 5.31$} &  {\scriptsize $49.05 ~ \pm ~ 1.56$} &  {\scriptsize $55.47 ~ \pm ~ 1.99$} &  {\scriptsize $55.34 ~ \pm ~ 3.05$} &  {\scriptsize $62.39 ~ \pm ~ 2.29$} &  {\scriptsize $67.59 ~ \pm ~ 2.36$} &  {\scriptsize $91.83 ~ \pm ~ 1.38$}\\[1ex]

    \hline
    \end{tabularx}%
    \caption{Few shot classification on the MSTAR dataset}
    \label{table:mstar}
\end{table*}

As shown in Fig.~\ref{fig:classif}, two different methods were employed to perform classification. The first method, `SAFE+k-NN`, involved using k-Nearest Neighbors (k-NN) on the extracted features, while the second method, `SAFE+linear`, applied linear regression with a cross-entropy loss. We compared our method against other feature extractors, a standard Principal Component Analysis (PCA) and a Uniform Manifold Approximation and Projection for Dimension Reduction (UMAP) \cite{mcinnes2018umap}, which is one of the most popular and effective dimensionality reduction algorithms. The data undergoes a zero mean and unit variance normalization before entering the feature extractors. For UMAP, we set the number of neighbours at 10, the minimum distance at 0.1 and 192 for the feature dimension, like a tiny ViT, while for the PCA the feature dimension is set at 50. We also compare our method with a ResNet18 classifier, aiming to replicate the network architecture described in \cite{inkawhich2021bridging}. This method was originally designed for classification on the SAMPLE dataset. In their experiments, classification was performed on thresholded PNG SAR images. To reproduce this, we will use amplitude images with a fixed threshold. Since the dataset is only composed of real images, adding Gaussian noise negatively affected the performance. However, we found that using label smoothing (lblsm) \cite{NEURIPS2019_f1748d6b} either maintained or slightly improved classification performance. The networks were trained until convergence, achieved after 150 epochs for the ResNet18 and 300 epochs for the linear regression model, using an Adam optimizer with a learning rate of $3 \times 10^{-3}$. We do not try to perform a linear evaluation on top of other feature extractors than SAFE. PCA's transformation depends on the entire dataset. As a result, the features of a single image can vary significantly based on the composition of the other images. Similar concerns apply to UMAP. It also constructs its embedding based on the relationships and distances between all the data points in the dataset.\\
       
\begin{figure}[t]
    \centering
    \includegraphics[width=\linewidth]{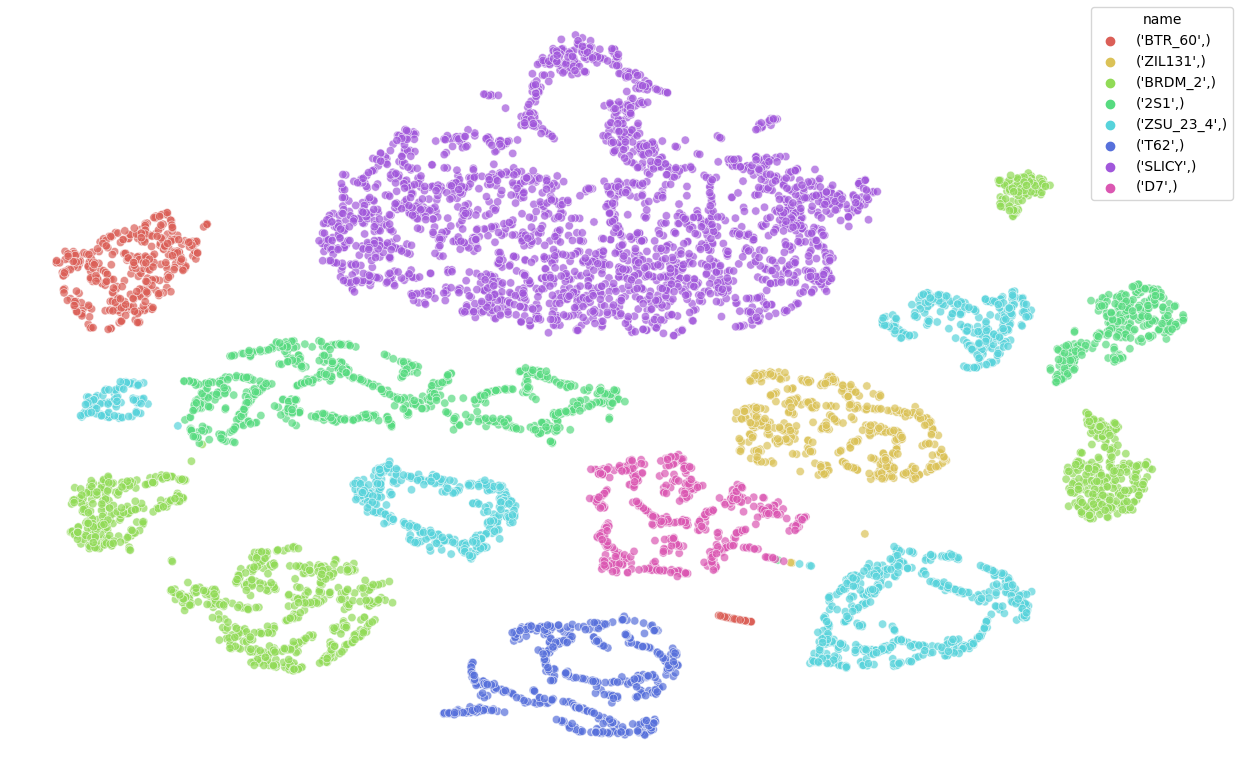}  
    \caption{t-SNE visualization of the MSTAR features extracted from the images.}
    \label{fig:mstar_tsne}
\end{figure}
Table~\ref{table:mstar} showcases the results of few-shot classification. When comparing the k-NN classification accuracy, SAFE reached the best results, getting $69\%$ with only one label per class and $94\%$ with ten. Using linear regression instead of k-NN yields even better results. The variance of the accuracy of our feature extractor is also smaller. It is significantly smaller than the one with ResNet, indicating that a pretrained frozen network results in more robust classification accuracy. ResNet does not perform well in few-shot learning scenarios; despite being a relatively small network with 11 million parameters, it requires more data to be effective. \\

This experiment demonstrates that our network can still extract discriminative features for specific classification tasks even when the training dataset differs from the evaluation dataset. The t-SNE \cite{van2008visualizing} visualization of the extracted features in Fig.~\ref{fig:mstar_tsne} shows that almost all points are grouped into clusters corresponding to their image classes, further illustrating the effectiveness of our feature extraction method on unseen data.

\subsection{Qualitative visualization}
\begin{figure*}[p]
\centering
\begin{minipage}[t]{.49\linewidth}
  \centering
  \begin{minipage}[b]{.32\linewidth}
      \centering
      \centerline{\includegraphics[width=2.9cm]{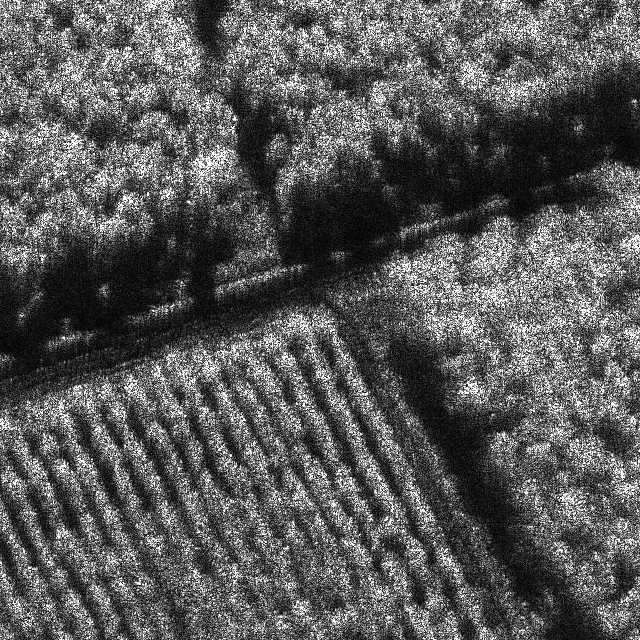}}
  \end{minipage}
  \hfill
  \begin{minipage}[b]{.32\linewidth}
      \centering
      \centerline{\includegraphics[width=2.9cm]{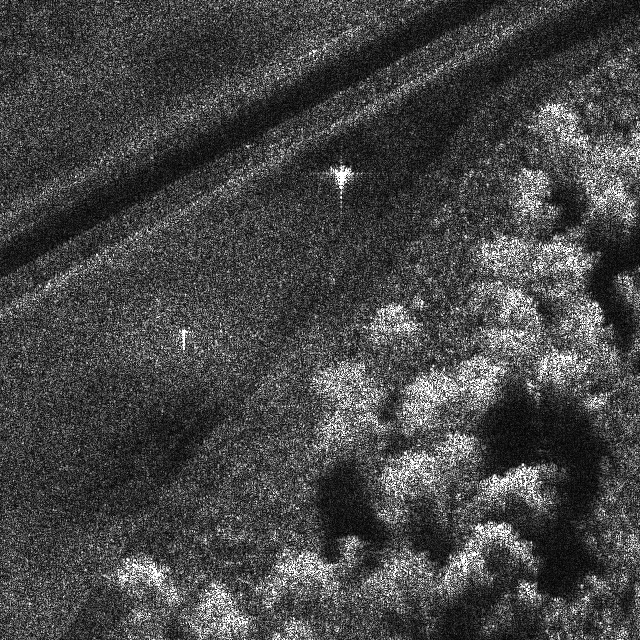}}
  \end{minipage}
  \hfill
  \begin{minipage}[b]{.32\linewidth}
      \centering
      \centerline{\includegraphics[width=2.9cm]{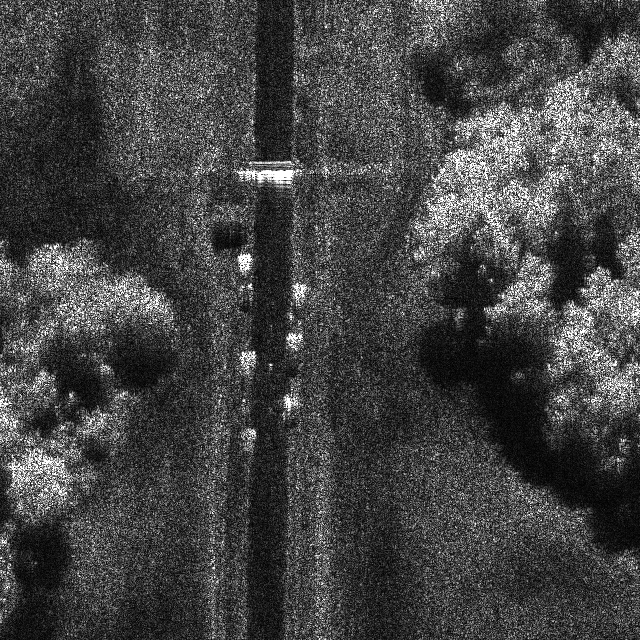}}
  \end{minipage}\\ \vspace{1mm}
  
  \begin{minipage}[b]{.32\linewidth}
      \centering
      \centerline{\includegraphics[width=2.9cm]{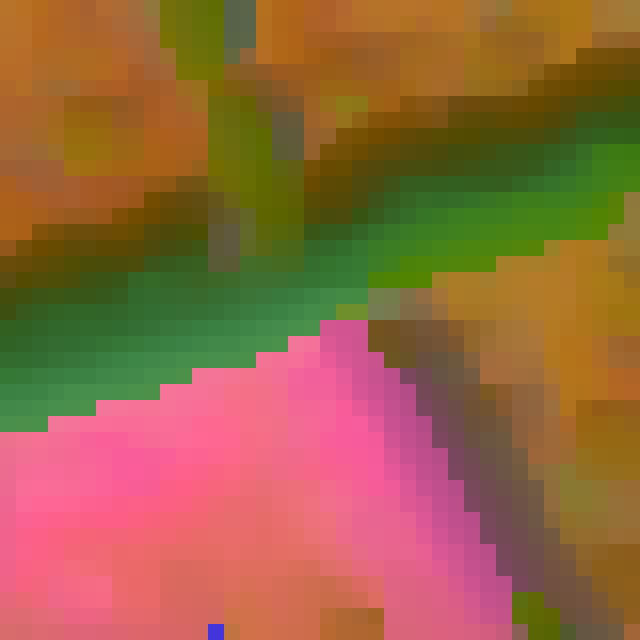}}
  \end{minipage}
  \hfill
  \begin{minipage}[b]{.32\linewidth}
      \centering
      \centerline{\includegraphics[width=2.9cm]{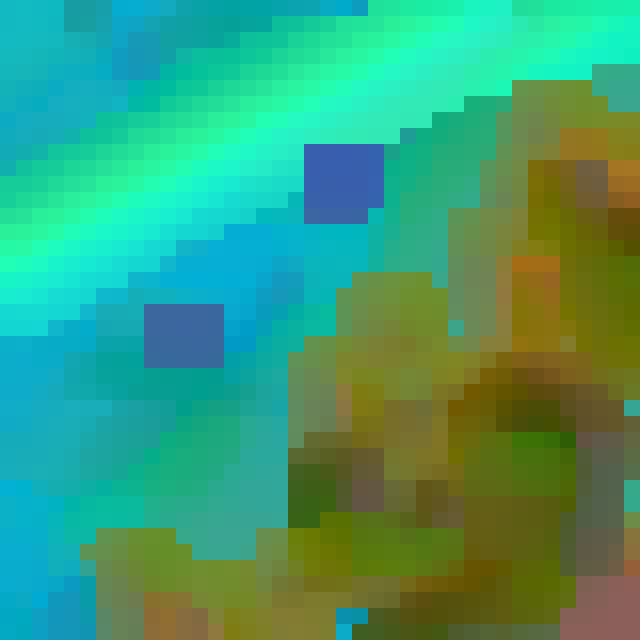}}
  \end{minipage}
  \hfill
  \begin{minipage}[b]{.32\linewidth}
      \centering
      \centerline{\includegraphics[width=2.9cm]{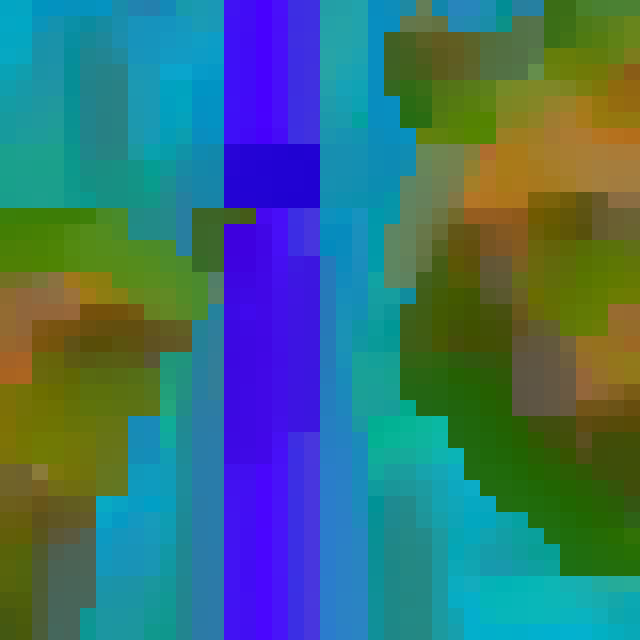}}
  \end{minipage}\\ \vspace{1mm}
  
  \begin{minipage}[b]{.32\linewidth}
      \centering
      \centerline{\includegraphics[width=2.9cm]{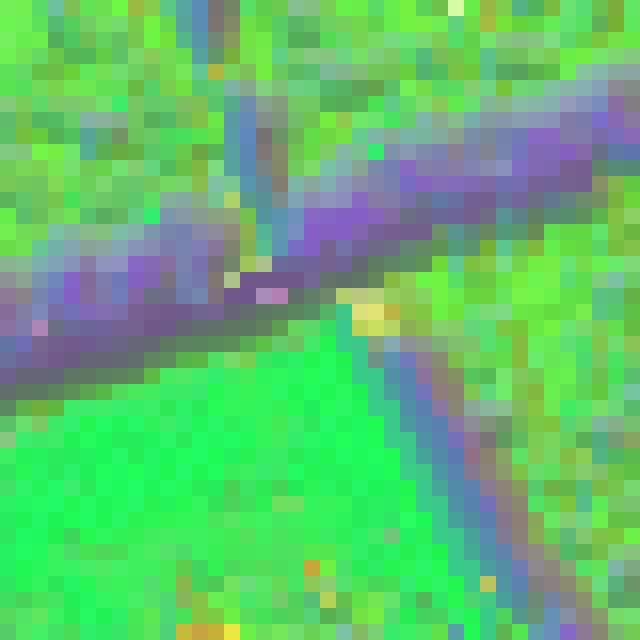}}
      \vspace{.1cm}
      \centerline{(1)}\medskip
  \end{minipage}
  \hfill
  \begin{minipage}[b]{.32\linewidth}
      \centering
      \centerline{\includegraphics[width=2.9cm]{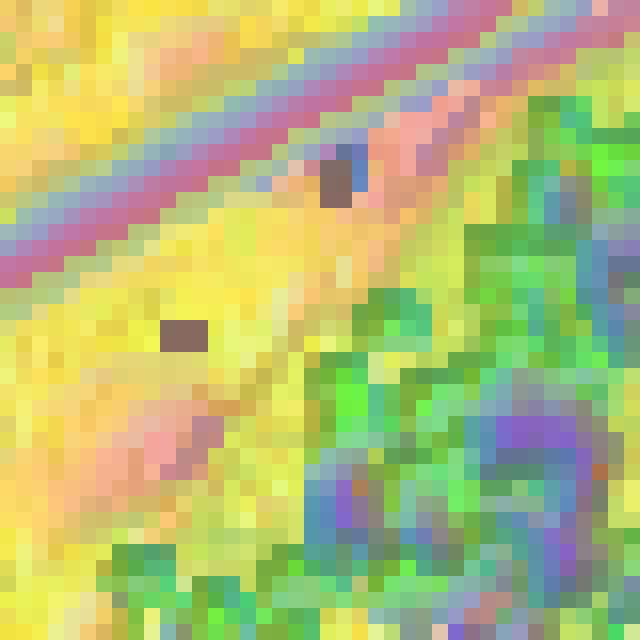}}
      \vspace{.1cm}
      \centerline{(2)}\medskip
  \end{minipage}
  \hfill
  \begin{minipage}[b]{.32\linewidth}
      \centering
      \centerline{\includegraphics[width=2.9cm]{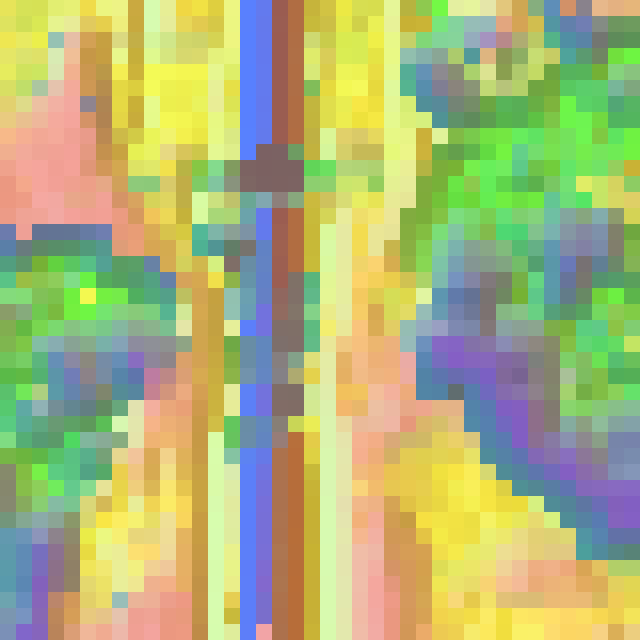}}
      \vspace{.1cm}
      \centerline{(3)}\medskip
  \end{minipage}
  \vspace{.1cm}
  \centerline{(a) SETHI X-band}\vspace{4mm}\centerline{}\medskip
\end{minipage}
\hfill
\begin{minipage}[t]{.49\linewidth}
  \centering
  \begin{minipage}[b]{.32\linewidth}
      \centering
      \centerline{\includegraphics[width=2.9cm]{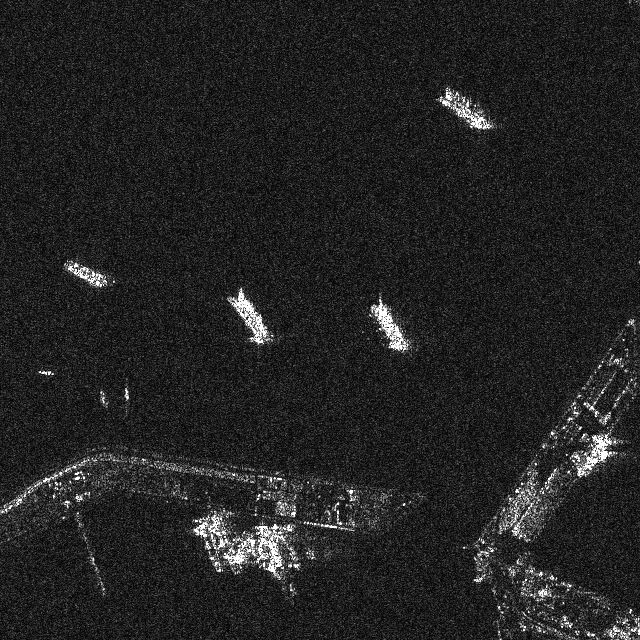}}
  \end{minipage}
  \hfill
  \begin{minipage}[b]{.32\linewidth}
      \centering
      \centerline{\includegraphics[width=2.9cm]{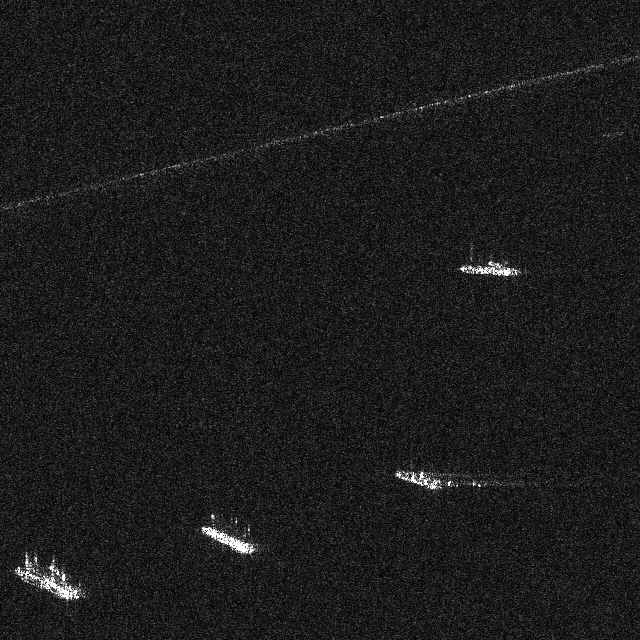}}
  \end{minipage}
  \hfill
  \begin{minipage}[b]{.32\linewidth}
      \centering
      \centerline{\includegraphics[width=2.9cm]{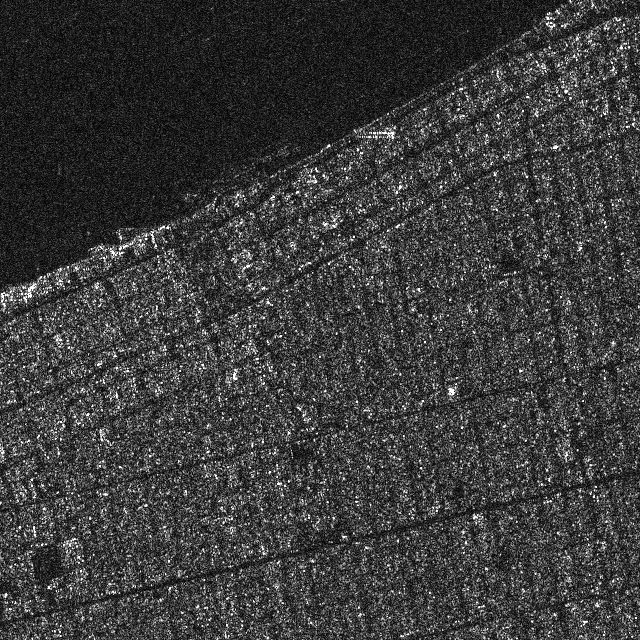}}
  \end{minipage}\\ \vspace{1mm}
  
  \begin{minipage}[b]{.32\linewidth}
      \centering
      \centerline{\includegraphics[width=2.9cm]{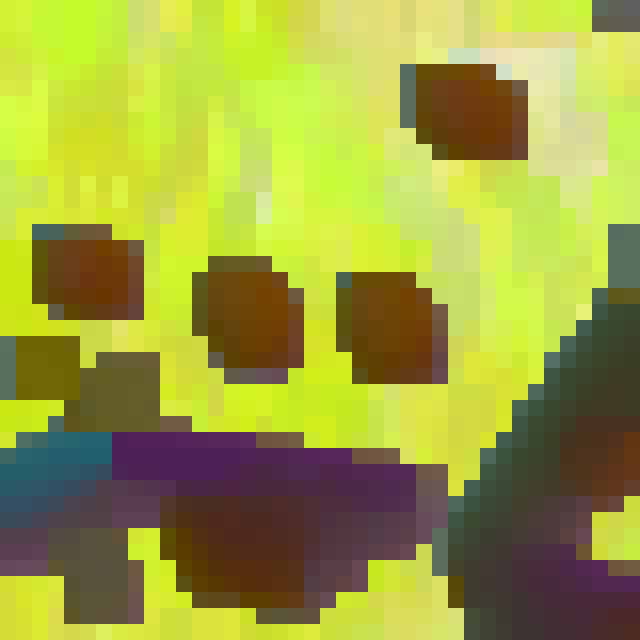}}
  \end{minipage}
  \hfill
  \begin{minipage}[b]{.32\linewidth}
      \centering
      \centerline{\includegraphics[width=2.9cm]{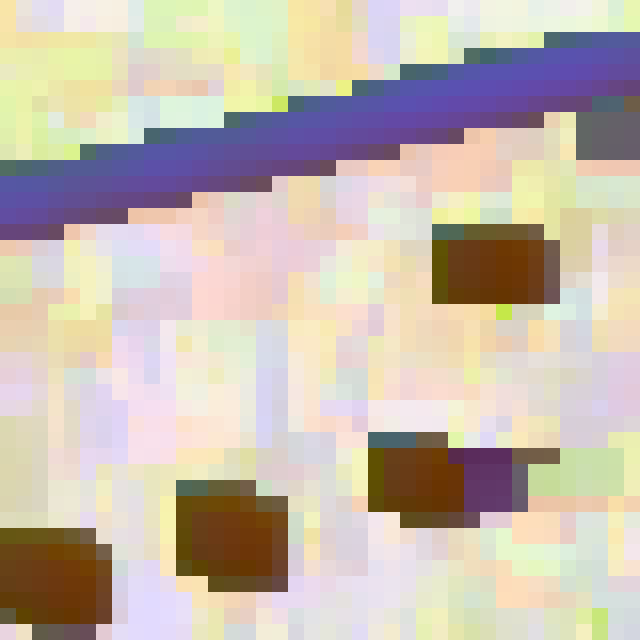}}
  \end{minipage}
  \hfill
  \begin{minipage}[b]{.32\linewidth}
      \centering
      \centerline{\includegraphics[width=2.9cm]{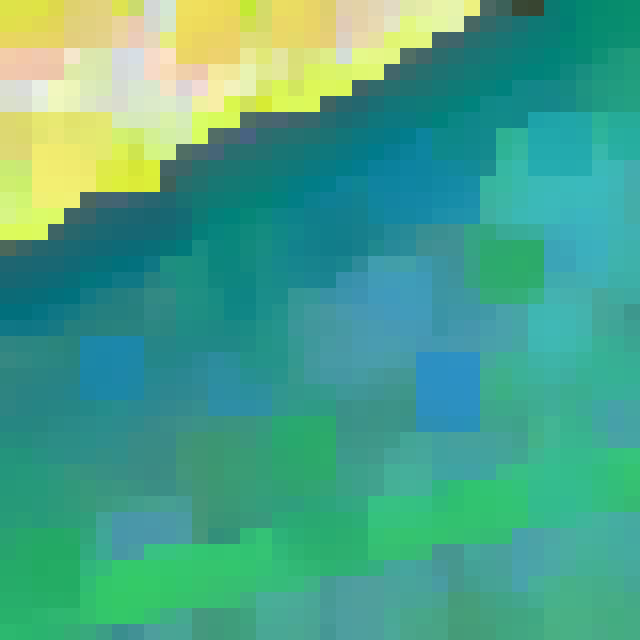}}
  \end{minipage}\\ \vspace{1mm}
  
  \begin{minipage}[b]{.32\linewidth}
      \centering
      \centerline{\includegraphics[width=2.9cm]{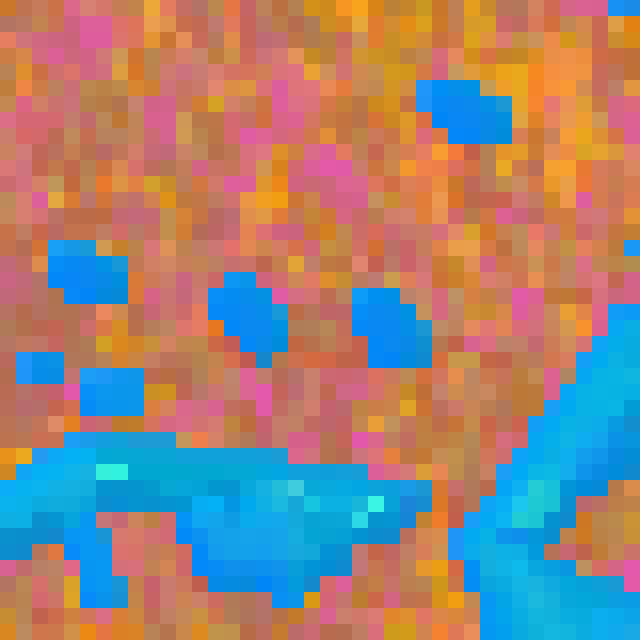}}
      \vspace{.1cm}
      \centerline{(1)}\medskip
  \end{minipage}
  \hfill
  \begin{minipage}[b]{.32\linewidth}
      \centering
      \centerline{\includegraphics[width=2.9cm]{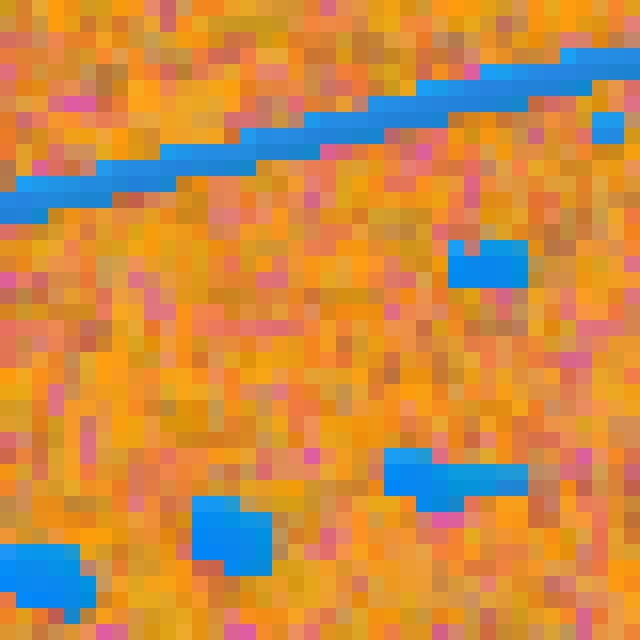}}
      \vspace{.1cm}
      \centerline{(2)}\medskip
  \end{minipage}
  \hfill
  \begin{minipage}[b]{.32\linewidth}
      \centering
      \centerline{\includegraphics[width=2.9cm]{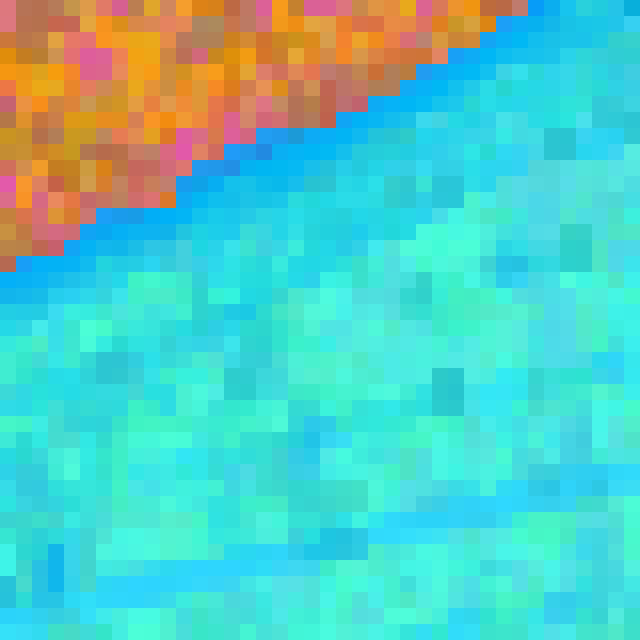}}
      \vspace{.1cm}
      \centerline{(3)}\medskip
  \end{minipage}
  \vspace{.1cm}
  \centerline{(b) Sentinel 1}\vspace{4mm}\centerline{}\medskip
\end{minipage}
\begin{minipage}[t]{.49\linewidth}
  \centering
  \begin{minipage}[b]{.32\linewidth}
      \centering
      \centerline{\includegraphics[width=2.9cm]{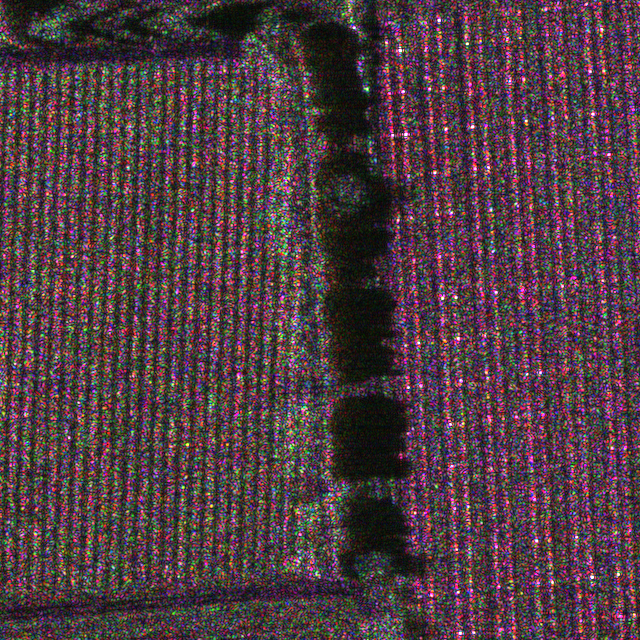}}
  \end{minipage}
  \hfill
  \begin{minipage}[b]{.32\linewidth}
      \centering
      \centerline{\includegraphics[width=2.9cm]{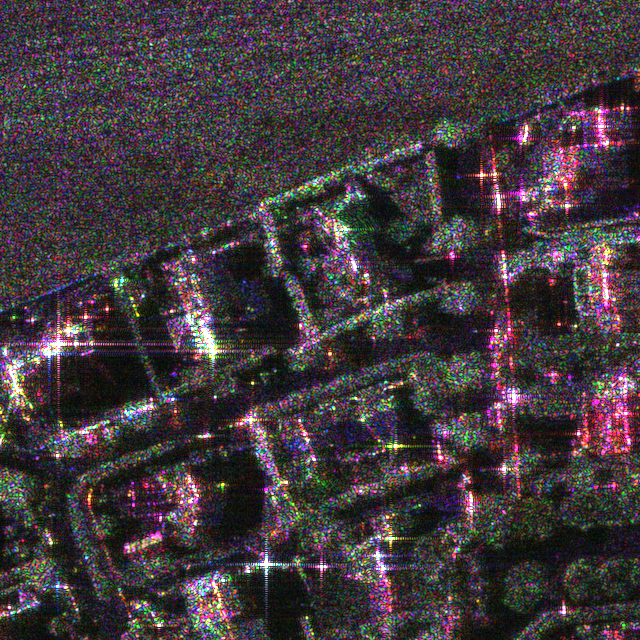}}
  \end{minipage}
  \hfill
  \begin{minipage}[b]{.32\linewidth}
      \centering
      \centerline{\includegraphics[width=2.9cm]{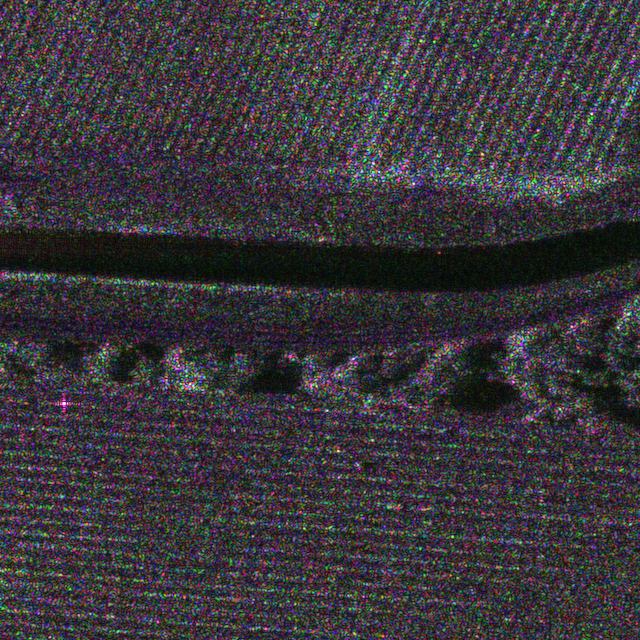}}
  \end{minipage}\\ \vspace{1mm}
  
  \begin{minipage}[b]{.32\linewidth}
      \centering
      \centerline{\includegraphics[width=2.9cm]{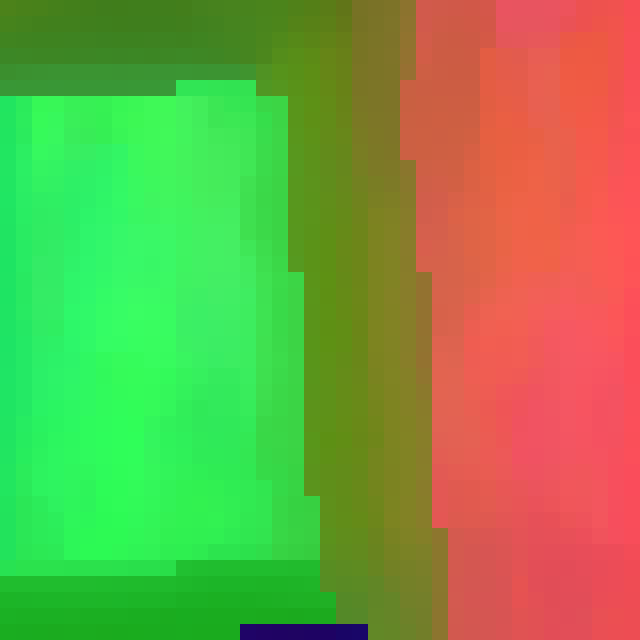}}
  \end{minipage}
  \hfill
  \begin{minipage}[b]{.32\linewidth}
      \centering
      \centerline{\includegraphics[width=2.9cm]{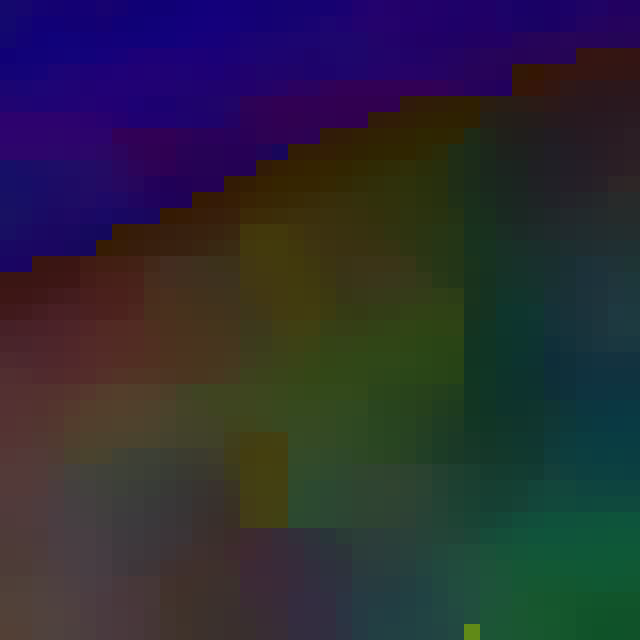}}
  \end{minipage}
  \hfill
  \begin{minipage}[b]{.32\linewidth}
      \centering
      \centerline{\includegraphics[width=2.9cm]{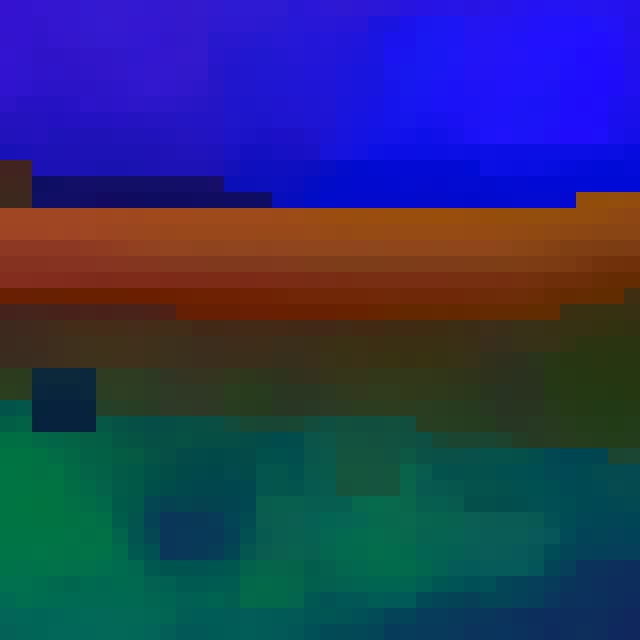}}
  \end{minipage}\\ \vspace{1mm}
  
  \begin{minipage}[b]{.32\linewidth}
      \centering
      \centerline{\includegraphics[width=2.9cm]{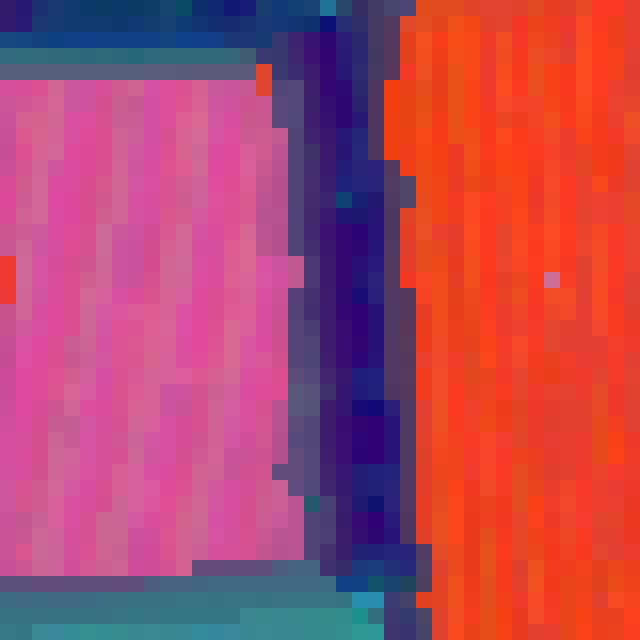}}
      \vspace{.1cm}
      \centerline{(1)}\medskip
  \end{minipage}
  \hfill
  \begin{minipage}[b]{.32\linewidth}
      \centering
      \centerline{\includegraphics[width=2.9cm]{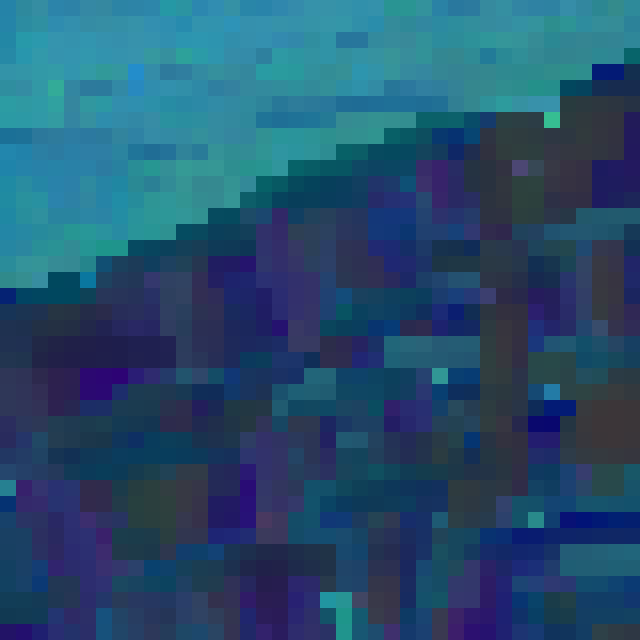}}
      \vspace{.1cm}
      \centerline{(2)}\medskip
  \end{minipage}
  \hfill
  \begin{minipage}[b]{.32\linewidth}
      \centering
      \centerline{\includegraphics[width=2.9cm]{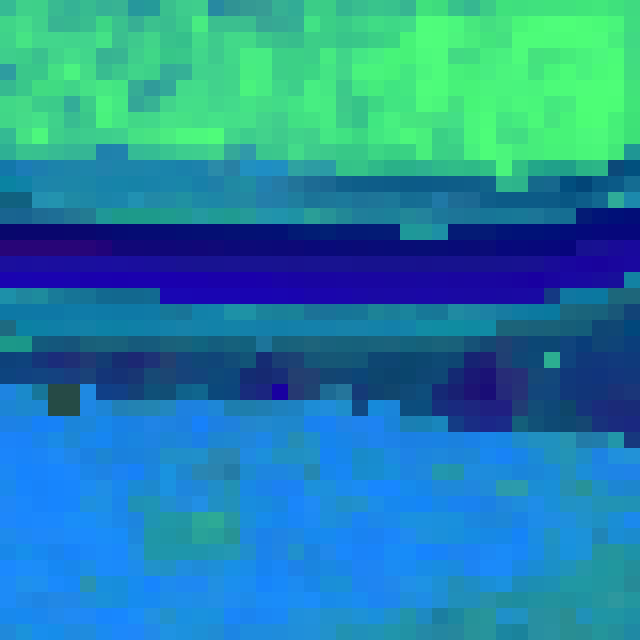}}
      \vspace{.1cm}
      \centerline{(3)}\medskip
  \end{minipage}
  \vspace{.1cm}
  \centerline{(c) PolSAR SETHI X-band}\vspace{0.1mm}\centerline{}\medskip
\end{minipage}
\hfill
\begin{minipage}[t]{.49\linewidth}
  \centering
  \begin{minipage}[b]{.32\linewidth}
      \centering
      \centerline{\includegraphics[width=2.9cm]{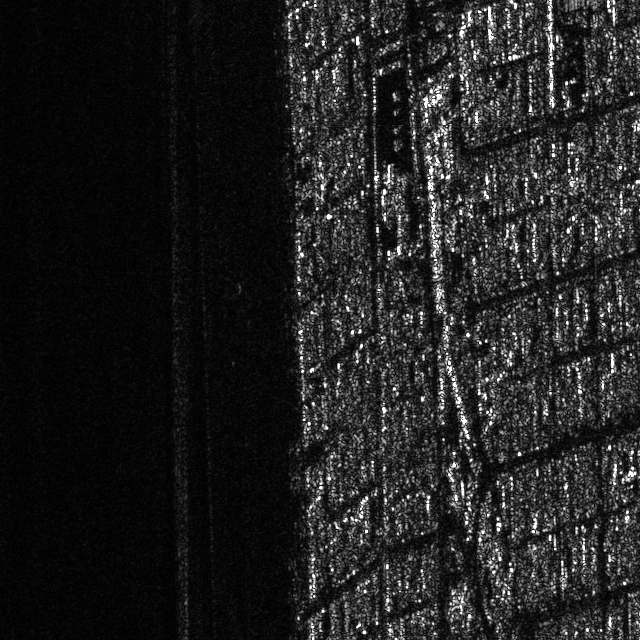}}
  \end{minipage}
  \hfill
  \begin{minipage}[b]{.32\linewidth}
      \centering
      \centerline{\includegraphics[width=2.9cm]{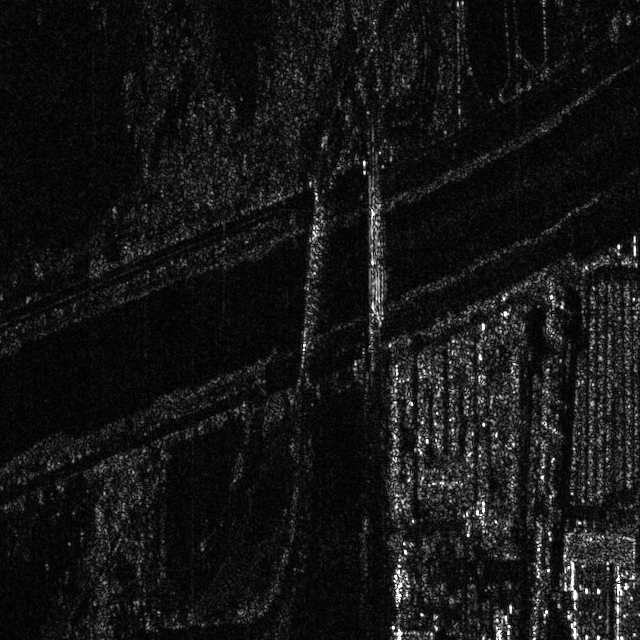}}
  \end{minipage}
  \hfill
  \begin{minipage}[b]{.32\linewidth}
      \centering
      \centerline{\includegraphics[width=2.9cm]{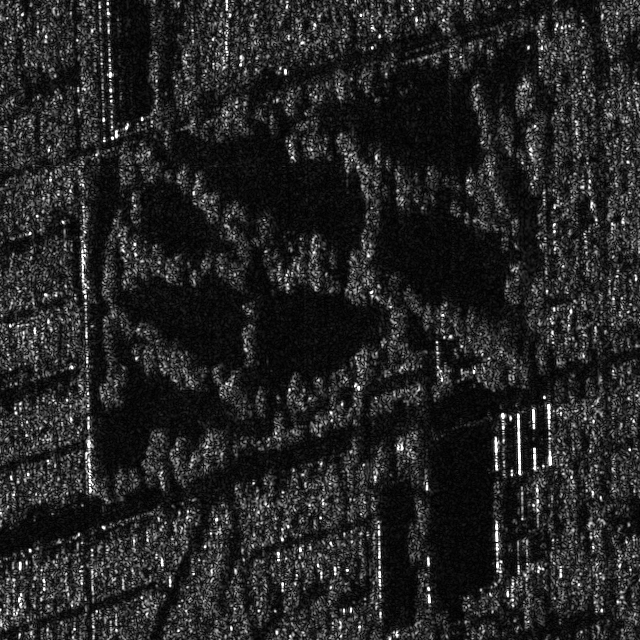}}
  \end{minipage}\\ \vspace{1mm}
  
  \begin{minipage}[b]{.32\linewidth}
      \centering
      \centerline{\includegraphics[width=2.9cm]{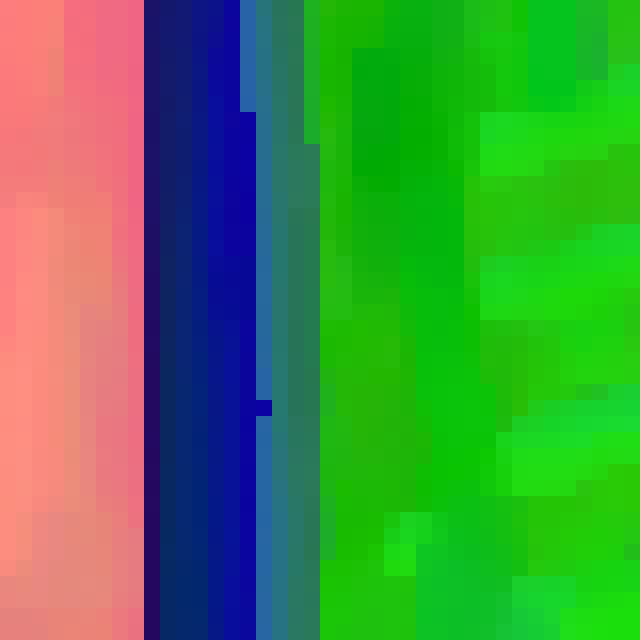}}
  \end{minipage}
  \hfill
  \begin{minipage}[b]{.32\linewidth}
      \centering
      \centerline{\includegraphics[width=2.9cm]{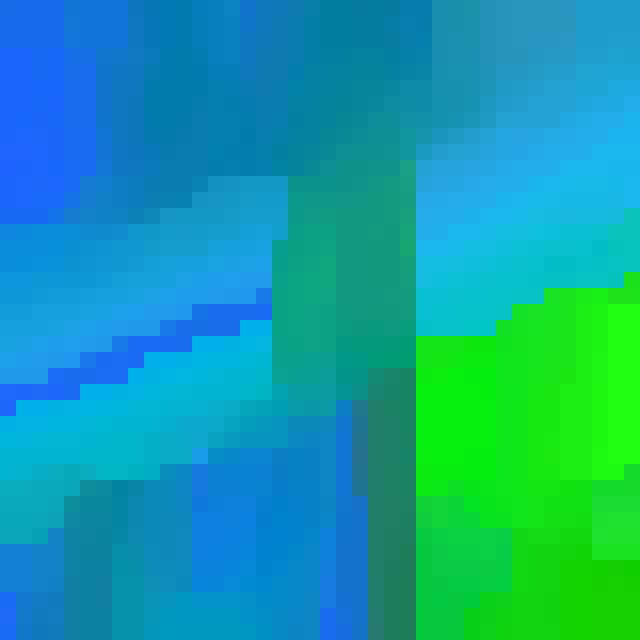}}
  \end{minipage}
  \hfill
  \begin{minipage}[b]{.32\linewidth}
      \centering
      \centerline{\includegraphics[width=2.9cm]{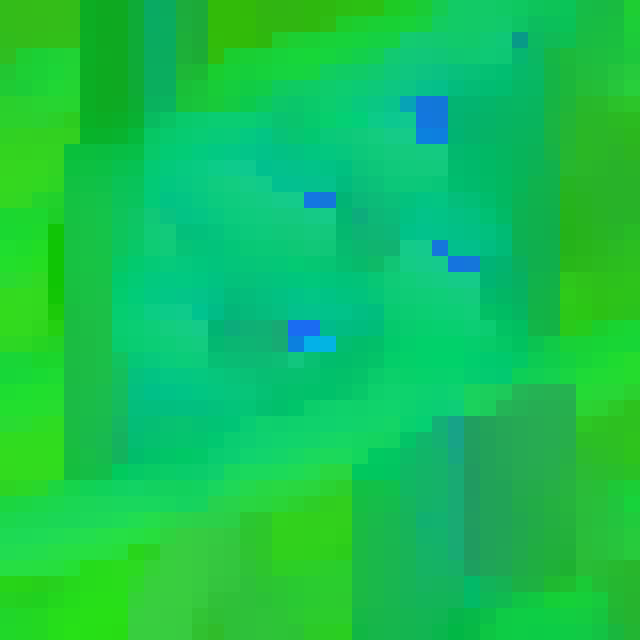}}
  \end{minipage}\\ \vspace{1mm}
  
  \begin{minipage}[b]{.32\linewidth}
      \centering
      \centerline{\includegraphics[width=2.9cm]{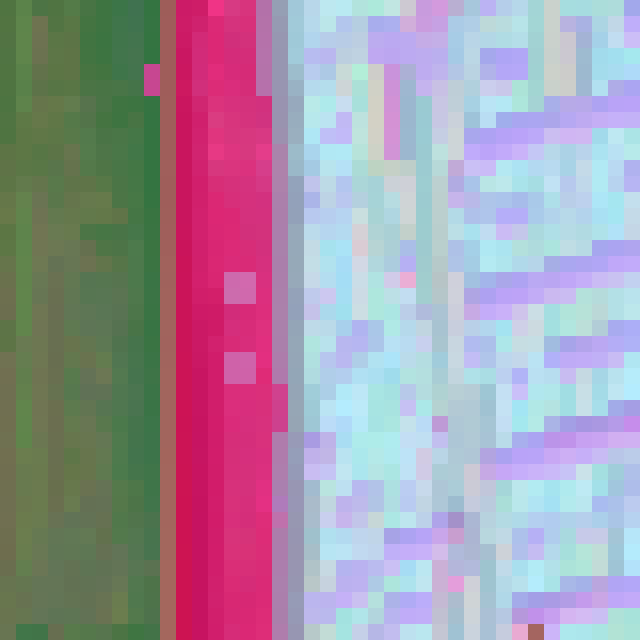}}
      \vspace{.1cm}
      \centerline{(1)}\medskip
  \end{minipage}
  \hfill
  \begin{minipage}[b]{.32\linewidth}
      \centering
      \centerline{\includegraphics[width=2.9cm]{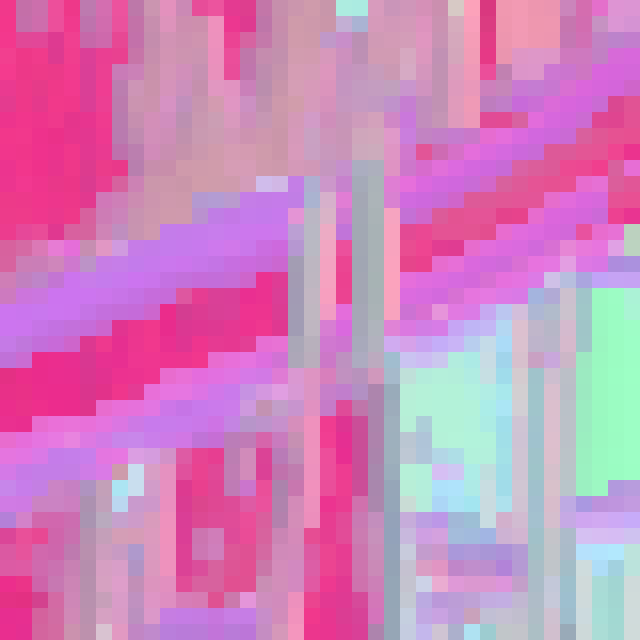}}
      \vspace{.1cm}
      \centerline{(2)}\medskip
  \end{minipage}
  \hfill
  \begin{minipage}[b]{.32\linewidth}
      \centering
      \centerline{\includegraphics[width=2.9cm]{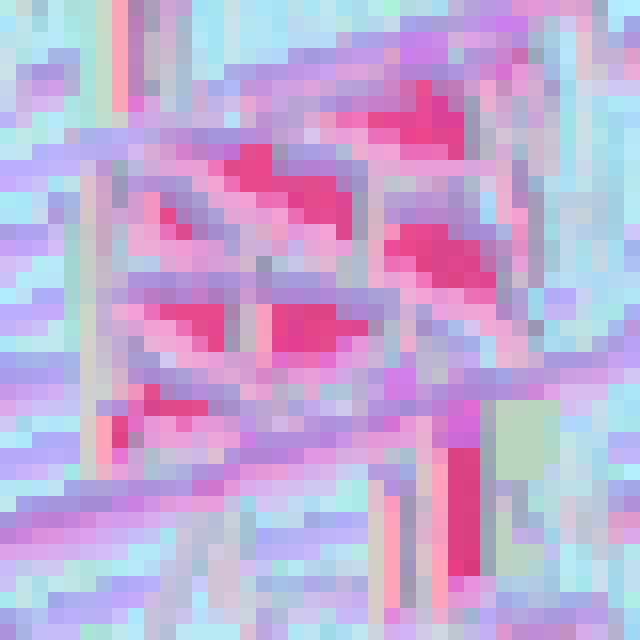}}
      \vspace{.1cm}
      \centerline{(3)}\medskip
  \end{minipage}
  \vspace{.1cm}
  \centerline{(d) UAVSAR}\vspace{0.1mm}\centerline{}\medskip
\end{minipage}

\caption{\centering Visualisation of the extracted features with SAFE. For each case, features are extracted with patches of size $64\times 64$ in the middle line and with $32\times 32$ patches in the bottom line. The dimension reduction algorithm UMAP is used on images grouped by columns (a), (b), (c) and (d).}
\label{fig:visu}
\end{figure*}
The quality of the extracted features can be evaluated visually. As for the segmentation evaluation, an image of shape $H\times W\times C$ is sliced in patches of size $P\times P$, and the patch window is shifted with a stride of $S$. Each patch is then processed by our feature extractor to obtain a matrix of shape $H/S\times W/S\times d_f$, $d_f$ being the dimension of the extracted features for each patch. After that, a dimension reduction algorithm is used to obtain a matrix of shape $H/S\times W/S\times 3$. The UMAP algorithm previously used for the classification tasks is used to reduce the dimension of the features. With a number of neighbors set at 15 and a minimum distance set at 0.1.

Some visualization examples are displayed in Fig.~\ref{fig:visu}. The stride is set to 16 with patches of size $64\times 64$ and $32\times 32$. It is important to note that the visual aspect of the features and the interpretation we derive from them depend heavily on the size of the patches.

For the $64\times 64$ patches, we observed that the meaning depends more on the spatial structure and local context than on the intensity of pixels. In the SETHI image (a)-(1), fields with similar intensity got a totally different feature representation caused by the difference in spatial patterns. In images (a)-(2) and (a)-(3), vehicles and fences, displayed as bright scatterers, have a feature representation closer to a road than to the forest, even though the forest had a higher intensity. This is likely because bright scatterers are often near low-reflectivity surfaces; for example, cars and buildings are often near or on roads. In the same images, we also see that the shadows caused by the three heights do not have a similar representation compared to the road, even though they both have a low reflectivity. Also, the shadows caused by the trees have a different representation compared to the road, even though both have low reflectivity. In Sentinel 1 images (b)-(1), (b)-(2), and (b)-(3), we observe similar results. Bright-intensity scatterers have a different representation. The ships can be differentiated from the shore, the line below the sea, and the city. Despite the different structures in the city, it is mapped to one cluster. We can also see a slightly brighter line at the bottom, corresponding to a wide road. In UAV images (d)-(1), (d)-(2) and (d)-(3) of Los Angeles, the ocean, the beach and the vegetation are clearly clustered. Interestingly, the network created a cluster of high-intensity and low-intensity structures nearby. It is the case for the limit between the beach and the city in (d)-(1), but also for the bridge in (d)-(2) and the golf course in the center of (d)-(3). We also notice that the water from the ocean in (d)-(1) and from Ballona Creek in (d)-(3) have completely different representations, surely due to the structures that are nearby.

For the $32\times 32$ patches, we observe that the extracted features rely more on the pixel intensity. In SETHI images (a)-(1),(a)-(2) and (a)-(3), all the vegetation is now mapped to the same color. The shadow caused by the trees also has a similar mapping to the road. However, the bright scatterers still have a feature representation closer to the low-reflectivity structures than the forest. For the Sentinel images (b)-(1), (b)-(2), and (b)-(3), there are now two distinct clusters: the sea and the man-made structures. The ships, the shore, and the city have a very similar representation, even though the city is a bit brighter. With smaller patches, we can also see smaller structures; in image (b)-(3), we can now distinguish smaller roads. As for the UAVSAR images in (d), the beach, Ballona Creek and some low-reflectivity parts of the vegetation and the golf course are now clustered together. We can also observe two new clusters, one for the roads and one for the trees. 

In Fig.~\ref{fig:visu} (c), a visualization is made on PolSAR images. The SAFE with four input channels is used to extract the features. We find that the network still accurately clusters the image by type of structure. Adding three other polarization channels makes the information richer, and thus, it is not as easy to group some similar structures that are not exactly the same. We also discarded Sentinel data and reduced the number of data by four in other cases ($VV$, $VH$, $HV$, and $HH$ are grouped in a single image). Combining these factors could explain why it is harder to understand high-level concepts. For example, the two fields in the image (c)-(1) are not gathered in the same cluster. We still see that the grass near the road and above the city, which has no field furrow, is clustered together. As previously, we observe that the road has similar features to bright points with $32\times 32$ patches.

\begin{figure}
\begin{minipage}[]{0.49\linewidth}
   \centering
   \includegraphics[width=\linewidth]{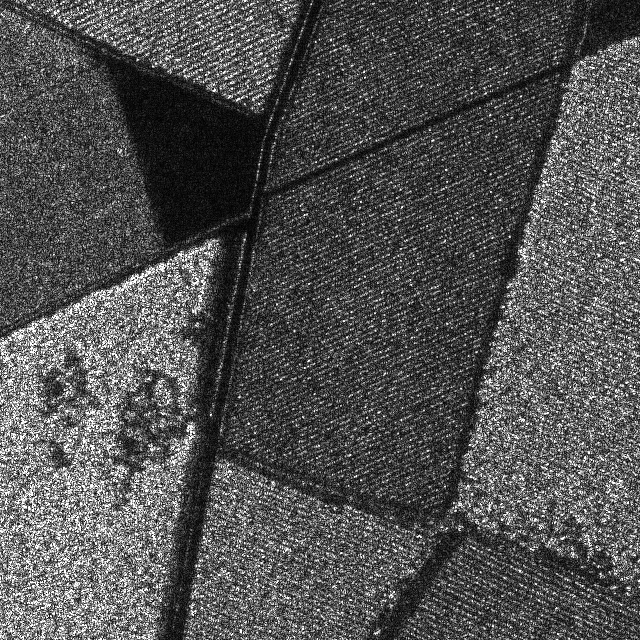}  
  \end{minipage}
    \hfill
  \begin{minipage}[]{0.49\linewidth}
   \centering
   \includegraphics[width=\linewidth]{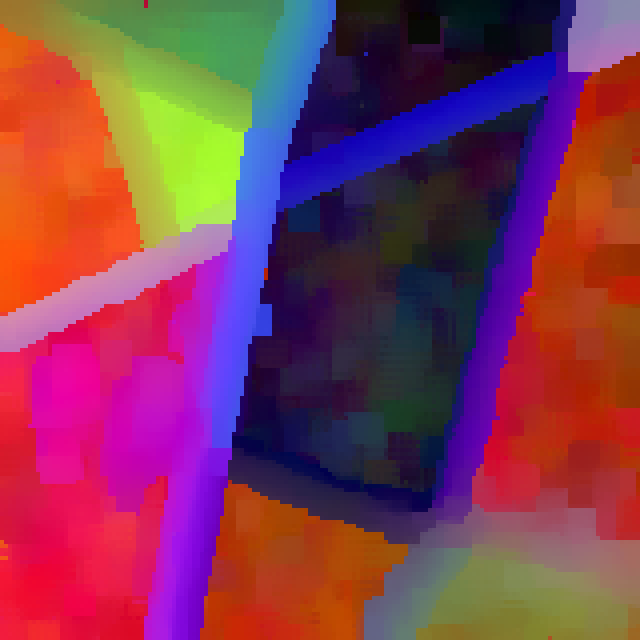}     
  \end{minipage}
  \caption{SLC image (left) and compressed features of $32\times 32$ with a stride of 4 (right). Features are extracted with SAFE and compressed with the UMAP algorithm.}
  \label{fig:visu_seg}
\end{figure}

A more refined clustering visualization is possible by taking only the area of interest. As displayed in Fig.~\ref{fig:visu_seg}, features of $32\times 32$ patches with a stride of 4 are visualized. It gives a representation that could approach segmentation models like Segment Anything \cite{kirillov2023segment} but trained for SAR images specifically.

\subsection{Pattern detection}

After demonstrating that the network can cluster different parts of the image using both intensity and local structure, we explore the possibility of recognizing a pattern type based on a single reference patch. Like the visualization process, the image is divided into patches, and the feature extractor processes each patch. A reference patch is also compressed into a vector. The goal is to compare the similarity of the image features with the reference features using cosine similarity, calculated as follows:
\begin{equation}
    c = \frac{ \left\langle f(\mathbf{X}), f(\mathbf{X}_{ref}) \right\rangle }{\left\|f(\mathbf{X})\right\|_2 \left\|f(\mathbf{X}_{ref})\right\|_2}
\label{eq:sim}
\end{equation}
where $\mathbf{X}$ is the image patch, $\mathbf{X}_{ref}$ is the reference patch and $f$ is the feature extractor.\\

\begin{figure*}[p]
\raggedright
\begin{minipage}[]{.03\linewidth}
   \centering
   \centerline{\rotatebox{90}{\small{Reference patch}} }  
\end{minipage}
\begin{minipage}[]{.155\linewidth}
    \centering
    \centerline{boat}\medskip
    \centerline{\includegraphics[width=2.8cm]{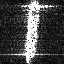}}
\end{minipage}
\begin{minipage}[]{.155\linewidth}
    \centering
    \centerline{city}\medskip
    \centerline{\includegraphics[width=2.8cm]{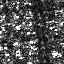}}
\end{minipage}
\begin{minipage}[]{.155\linewidth}
    \centering
    \centerline{urban}\medskip
    \centerline{\includegraphics[width=2.8cm]{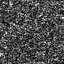}}
\end{minipage}\\ \vspace{1mm}

\begin{minipage}[]{.03\linewidth}
    \centering
    \rotatebox{90}{\small{Test patches}}
\end{minipage}
\begin{minipage}[]{.155\linewidth}
    \centering
    \centerline{\includegraphics[width=2.8cm]{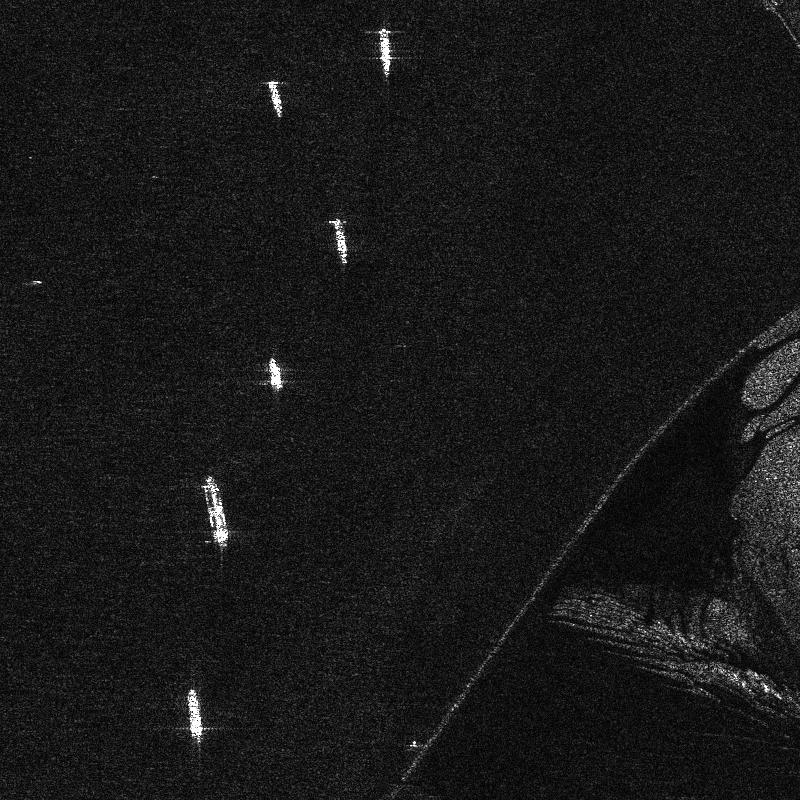}}
\end{minipage}
\begin{minipage}[]{.155\linewidth}
    \centering
    \centerline{\includegraphics[width=2.8cm]{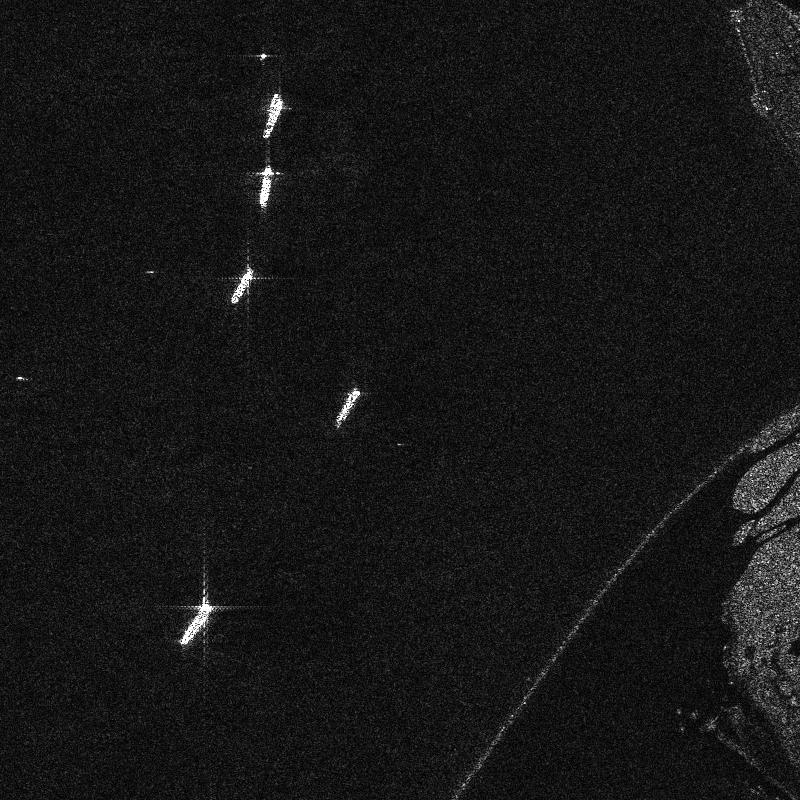}}
\end{minipage}
\begin{minipage}[]{.155\linewidth}
    \centering
    \centerline{\includegraphics[width=2.8cm]{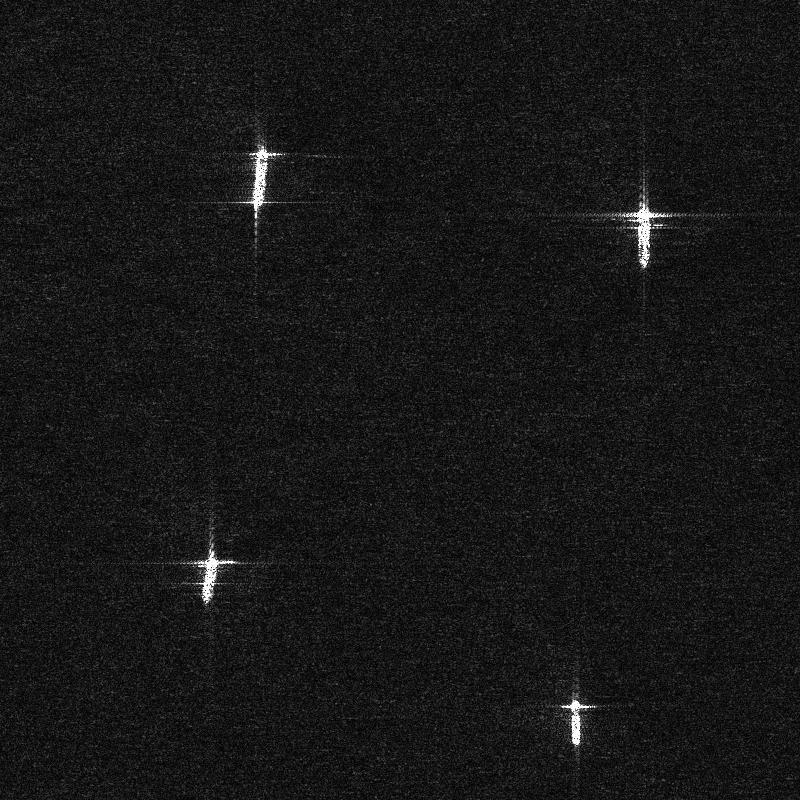}}
\end{minipage}
\begin{minipage}[]{.155\linewidth}
    \centering
    \centerline{\includegraphics[width=2.8cm]{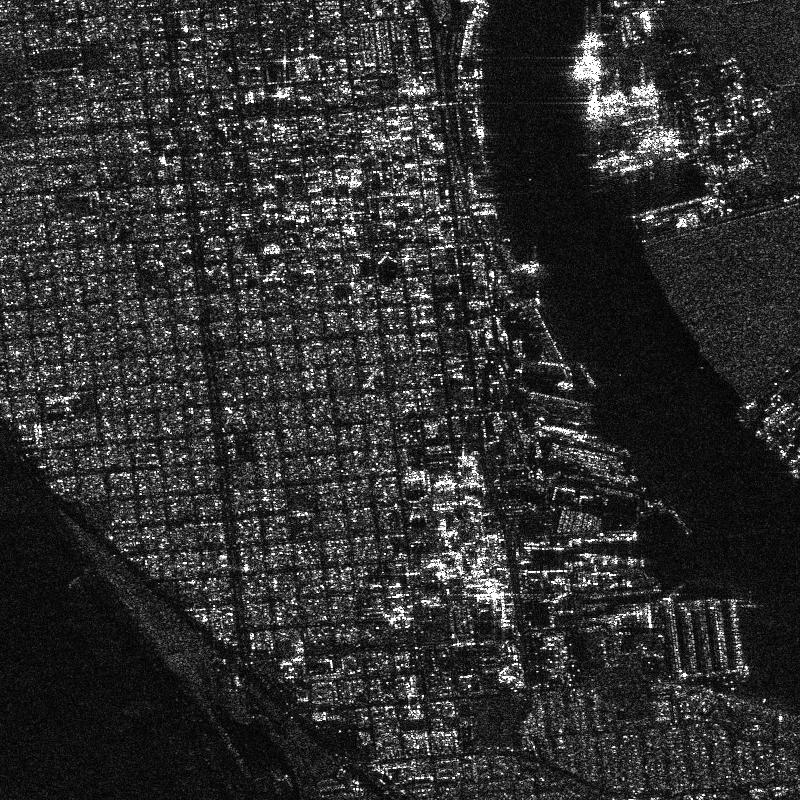}}
\end{minipage}
\begin{minipage}[]{.155\linewidth}
    \centering
    \centerline{\includegraphics[width=2.8cm]{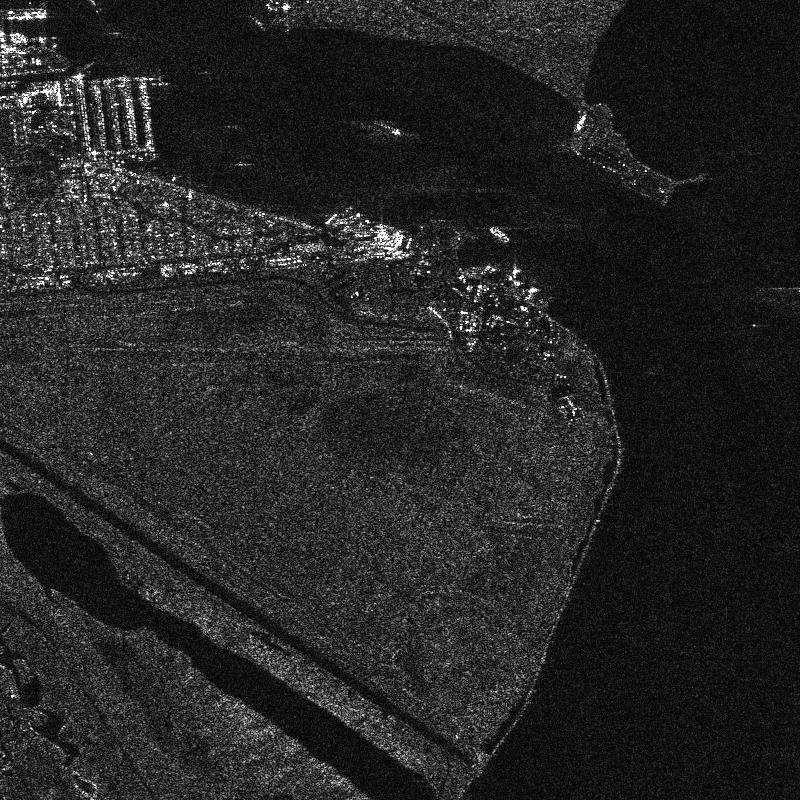}}
\end{minipage}
\begin{minipage}[]{.155\linewidth}
    \centering
    \centerline{\includegraphics[width=2.8cm]{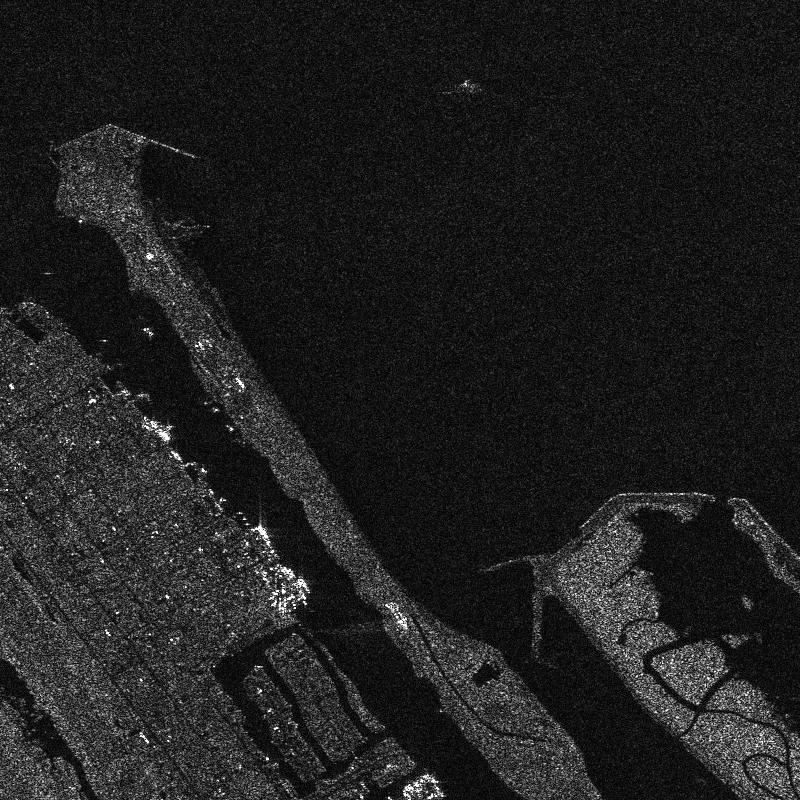}}
\end{minipage}\\ \vspace{1mm}

\begin{minipage}[]{.03\linewidth}
    \centering
    \rotatebox{90}{\small{``boat" similarity}}
\end{minipage}
\begin{minipage}[]{.155\linewidth}
    \centering
    \centerline{\includegraphics[width=2.8cm]{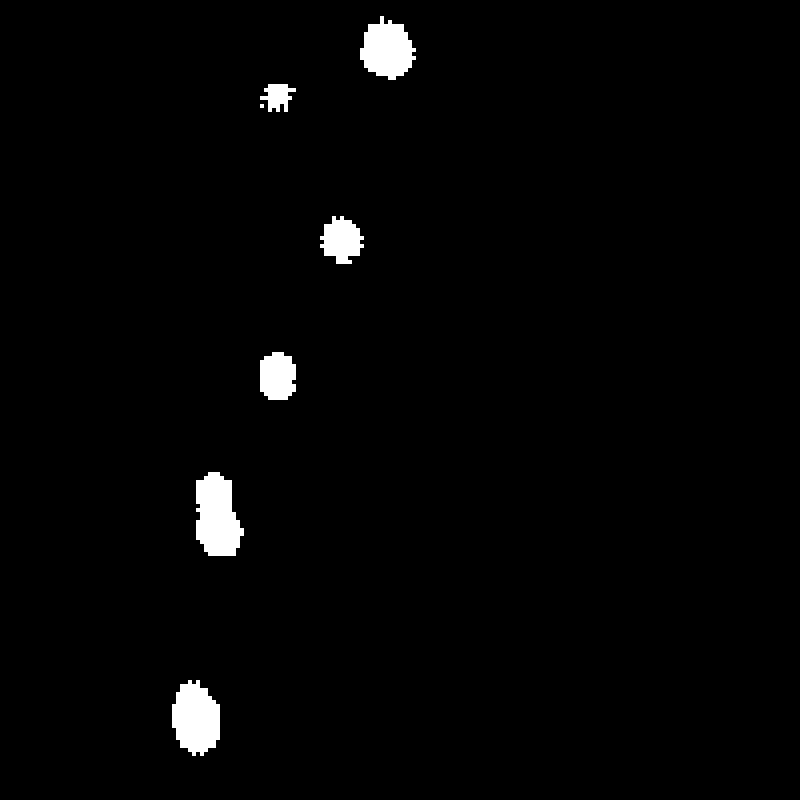}}
\end{minipage}
\begin{minipage}[]{.155\linewidth}
    \centering
    \centerline{\includegraphics[width=2.8cm]{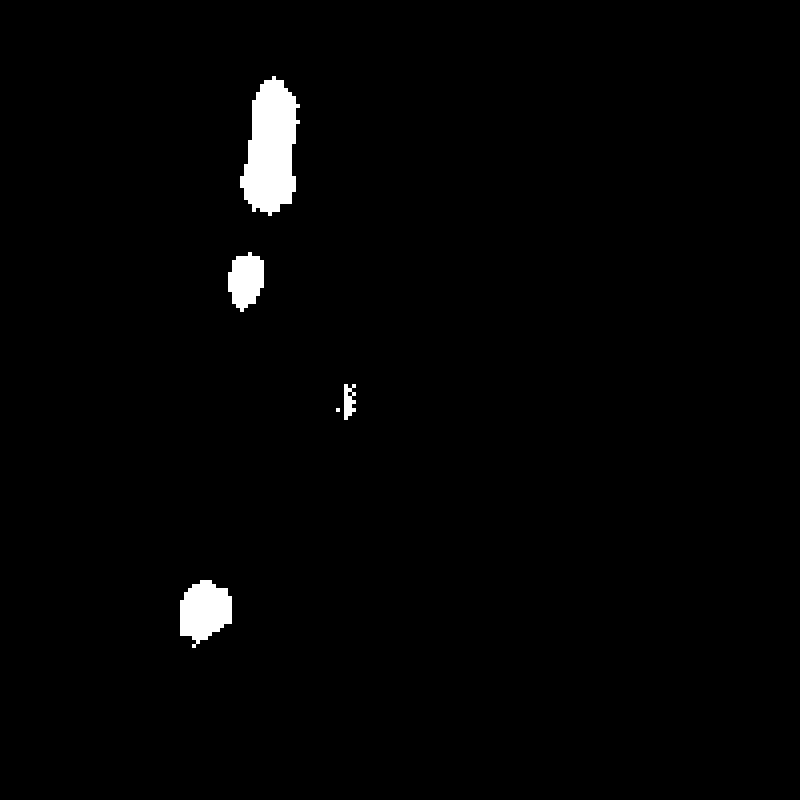}}
\end{minipage}
\begin{minipage}[]{.155\linewidth}
    \centering
    \centerline{\includegraphics[width=2.8cm]{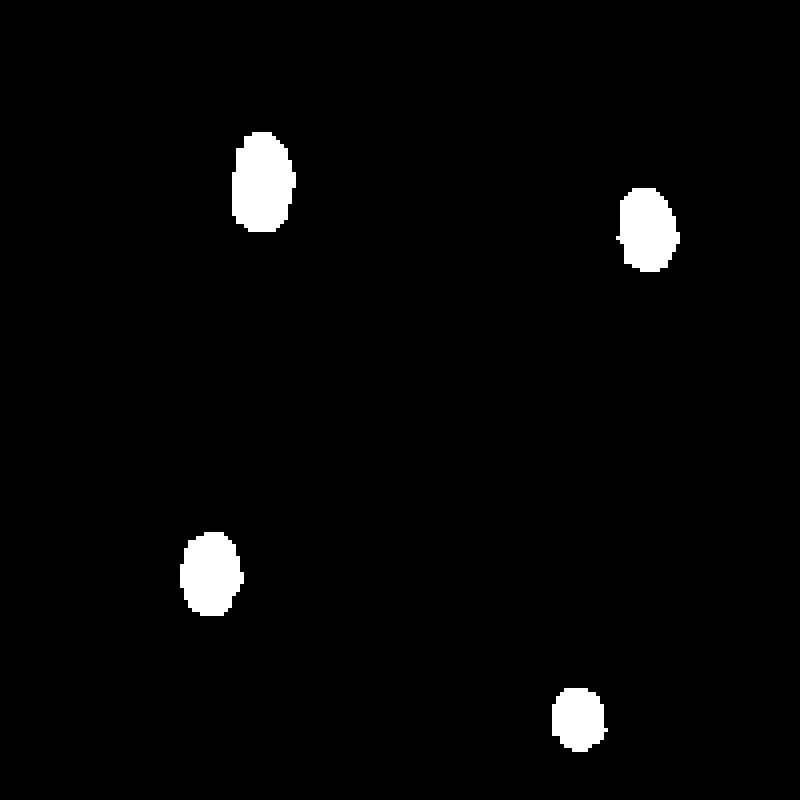}}
\end{minipage}
\begin{minipage}[]{.155\linewidth}
    \centering
    \centerline{\includegraphics[width=2.8cm]{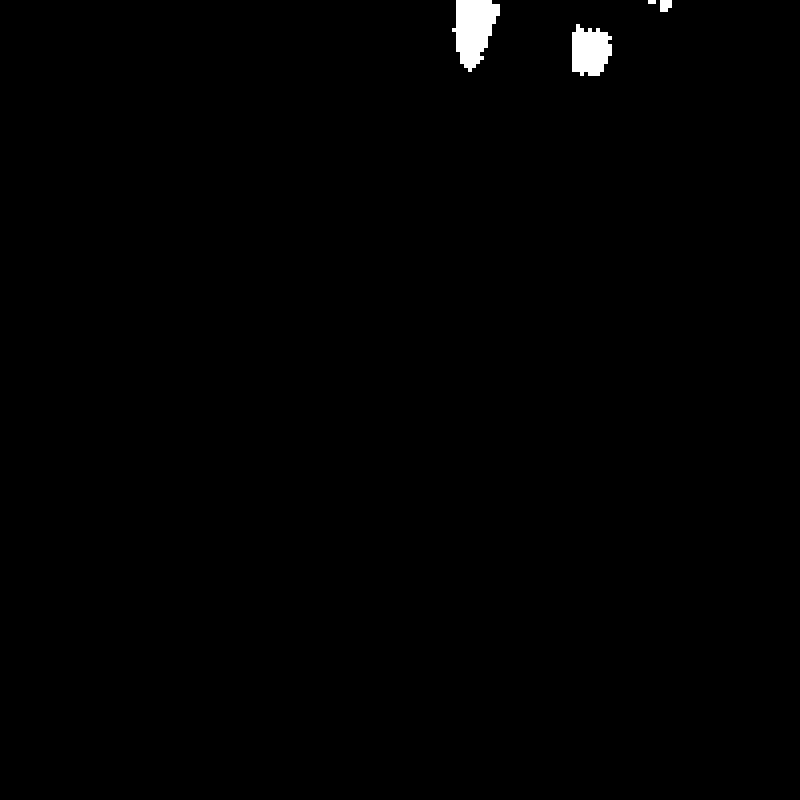}}
\end{minipage}
\begin{minipage}[]{.155\linewidth}
    \centering
    \centerline{\includegraphics[width=2.8cm]{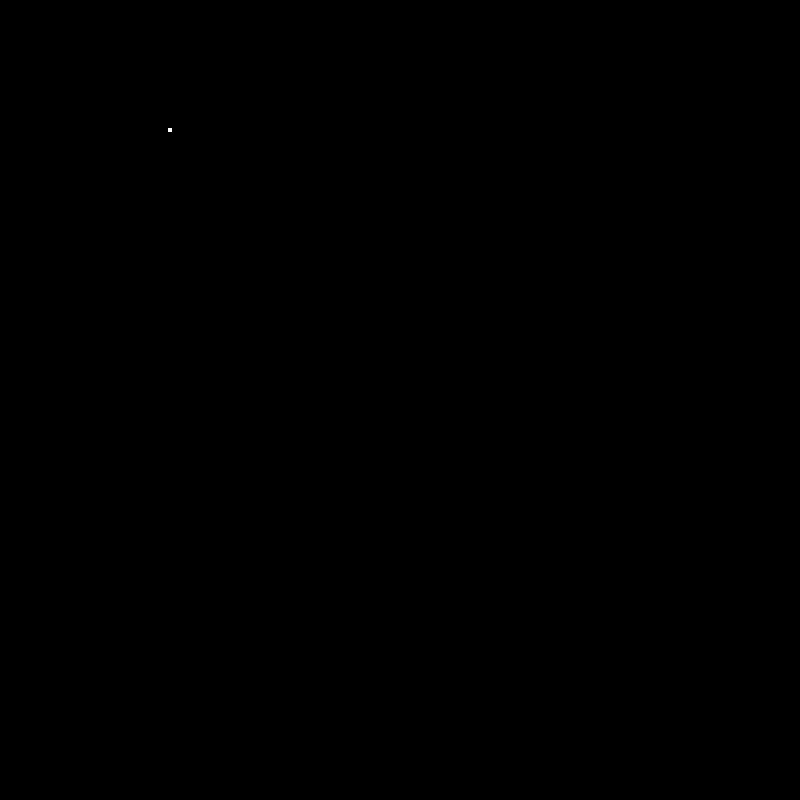}}
\end{minipage}
\begin{minipage}[]{.155\linewidth}
    \centering
    \centerline{\includegraphics[width=2.8cm]{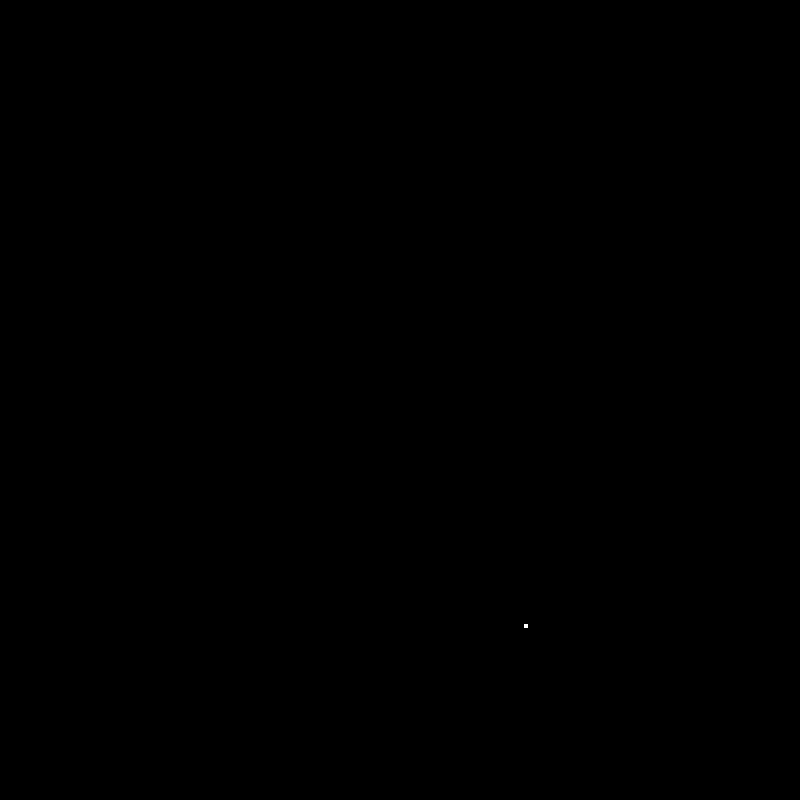}}
\end{minipage}\\ \vspace{1mm}

\begin{minipage}[]{.03\linewidth}
    \centering
    \rotatebox{90}{\small{``city" similarity}}
\end{minipage}
\begin{minipage}[]{.155\linewidth}
    \centering
    \centerline{\includegraphics[width=2.8cm]{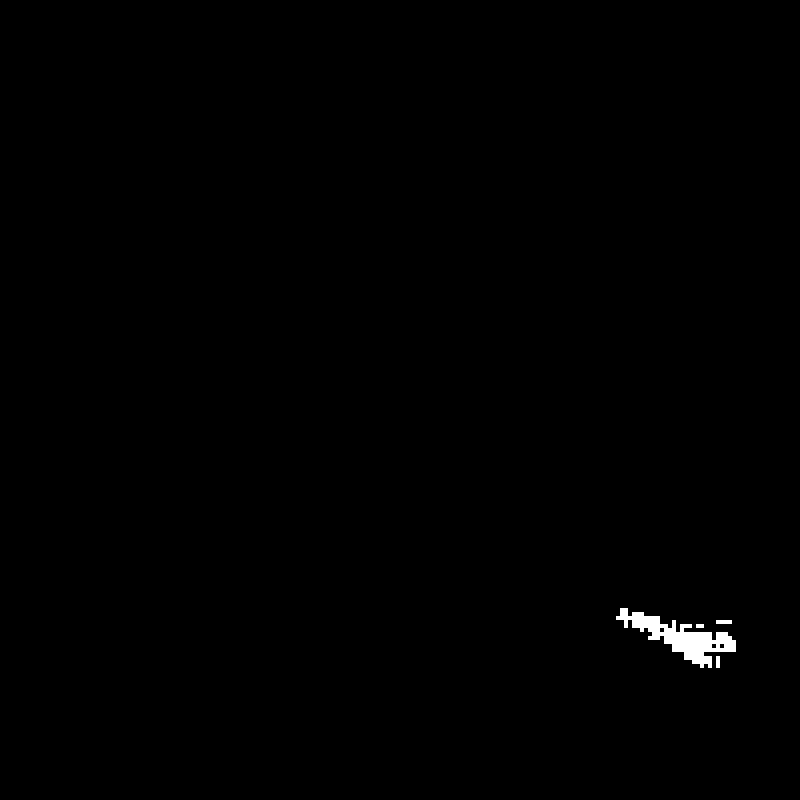}}
\end{minipage}
\begin{minipage}[]{.155\linewidth}
    \centering
    \centerline{\includegraphics[width=2.8cm]{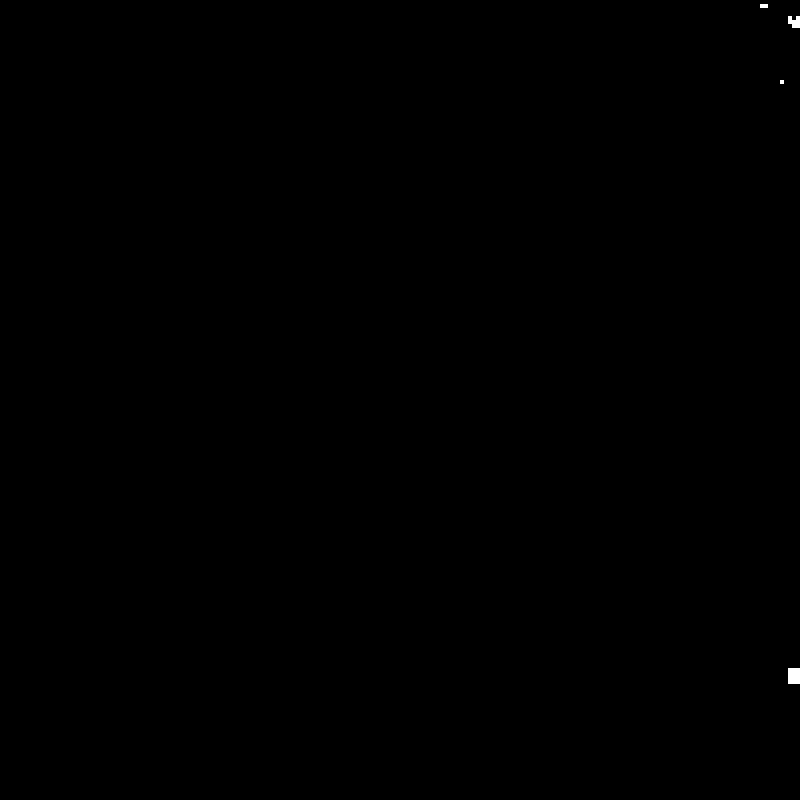}}
\end{minipage}
\begin{minipage}[]{.155\linewidth}
    \centering
    \centerline{\includegraphics[width=2.8cm]{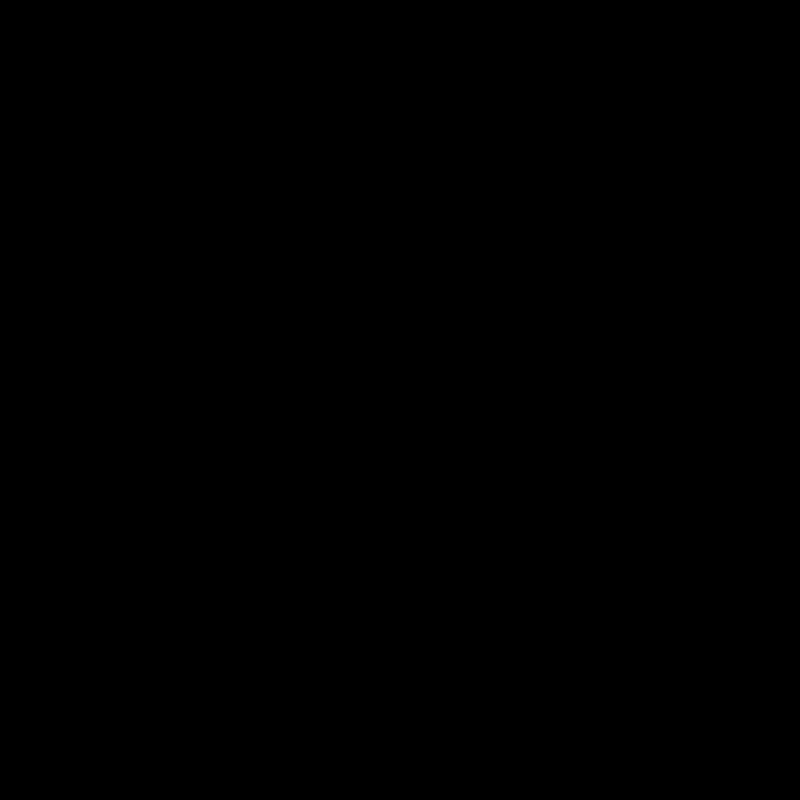}}
\end{minipage}
\begin{minipage}[]{.155\linewidth}
    \centering
    \centerline{\includegraphics[width=2.8cm]{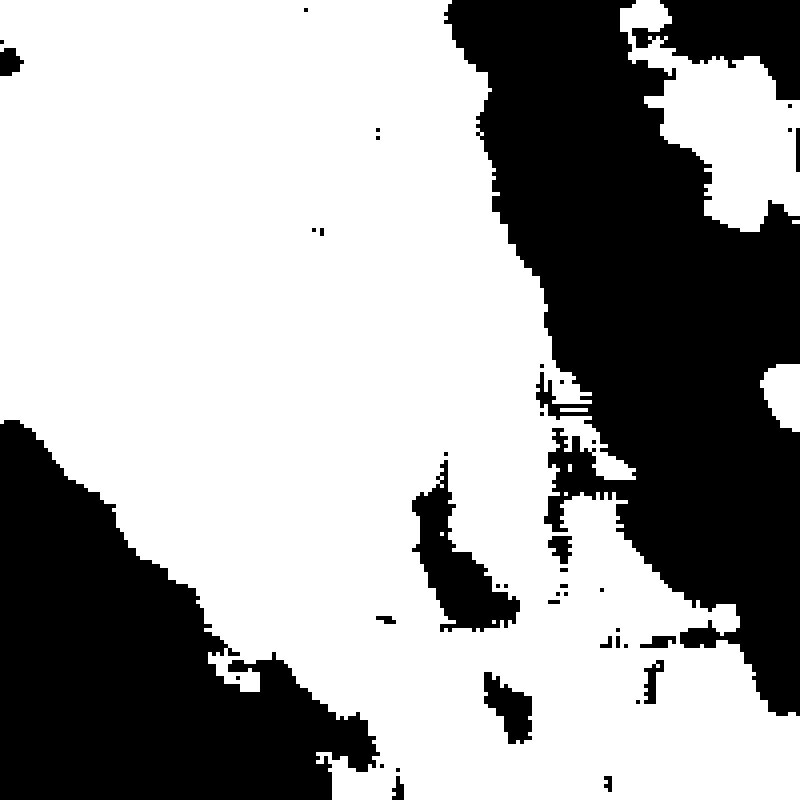}}
\end{minipage}
\begin{minipage}[]{.155\linewidth}
    \centering
    \centerline{\includegraphics[width=2.8cm]{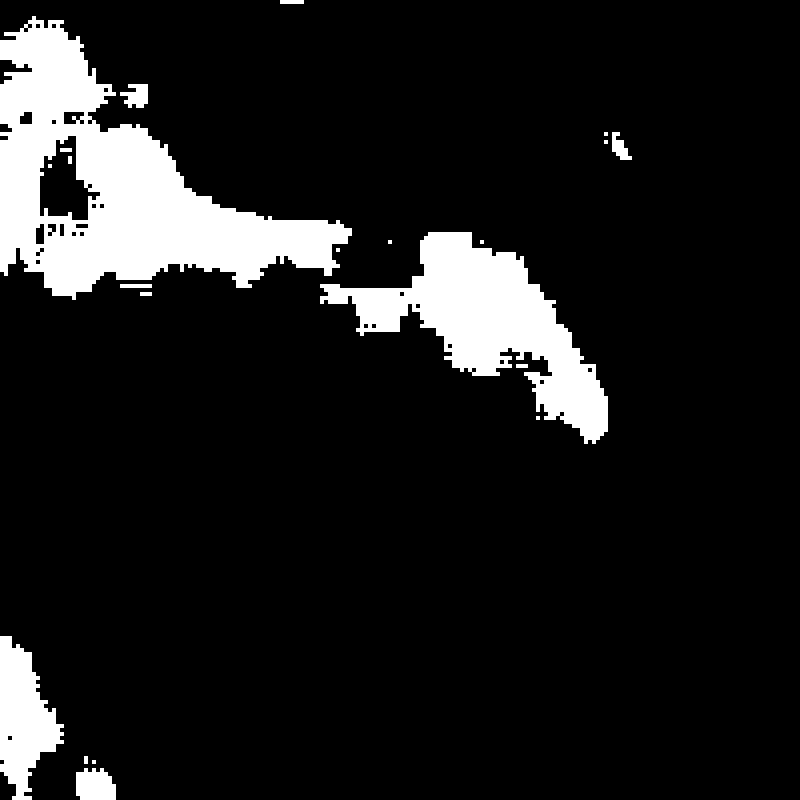}}
\end{minipage}
\begin{minipage}[]{.155\linewidth}
    \centering
    \centerline{\includegraphics[width=2.8cm]{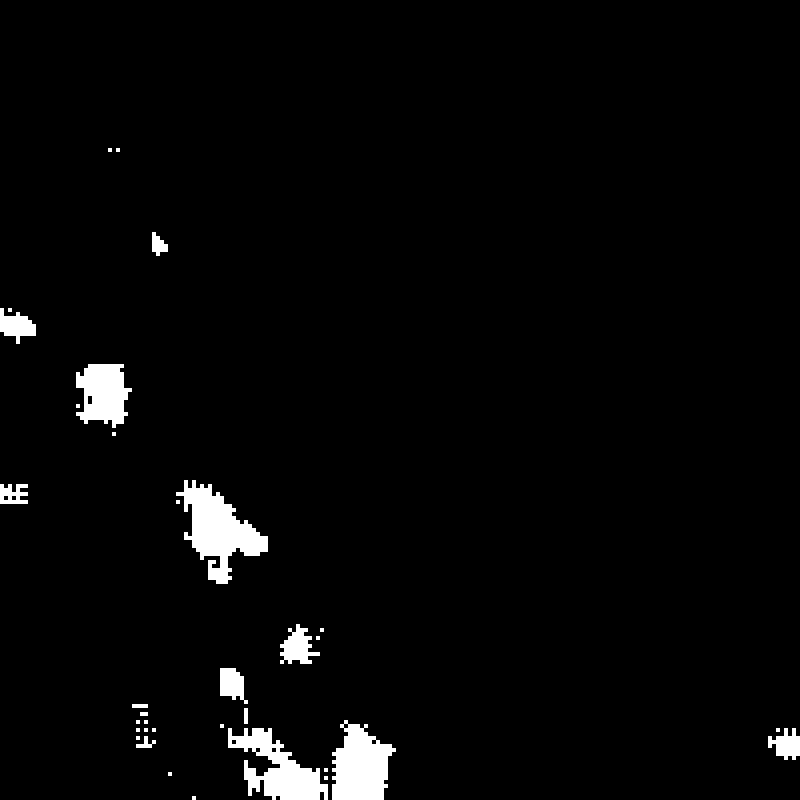}}
\end{minipage}\\ \vspace{1mm}

\begin{minipage}[]{.03\linewidth}
    \centering
    \rotatebox{90}{\small{``urban" similarity}}
\end{minipage}
\begin{minipage}[]{.155\linewidth}
    \centering
    \centerline{\includegraphics[width=2.8cm]{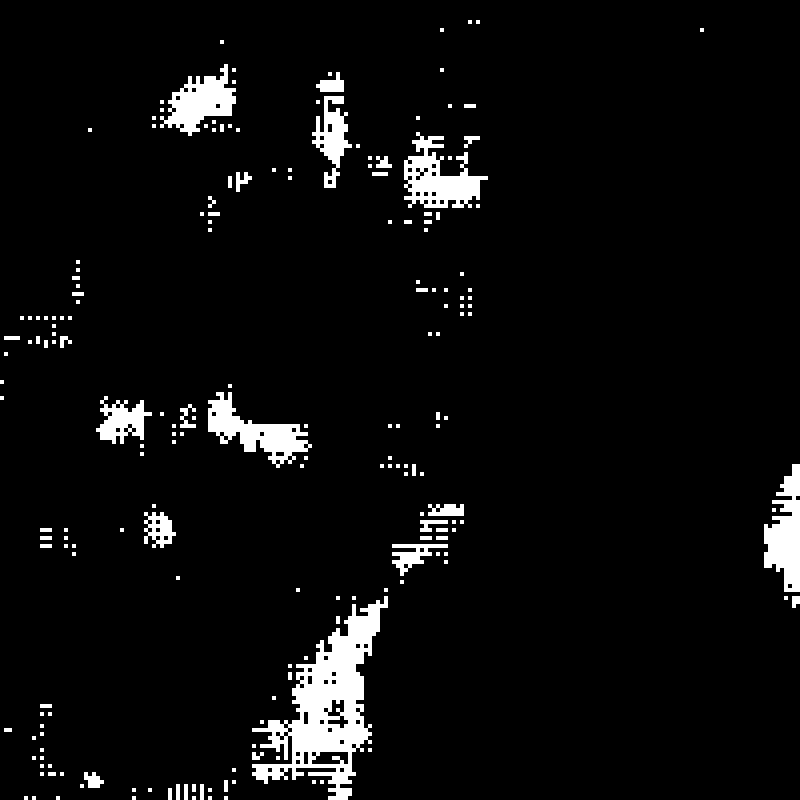}}
     \vspace{.1cm}
      \centerline{(1)}\medskip
\end{minipage}
\begin{minipage}[]{.155\linewidth}
    \centering
    \centerline{\includegraphics[width=2.8cm]{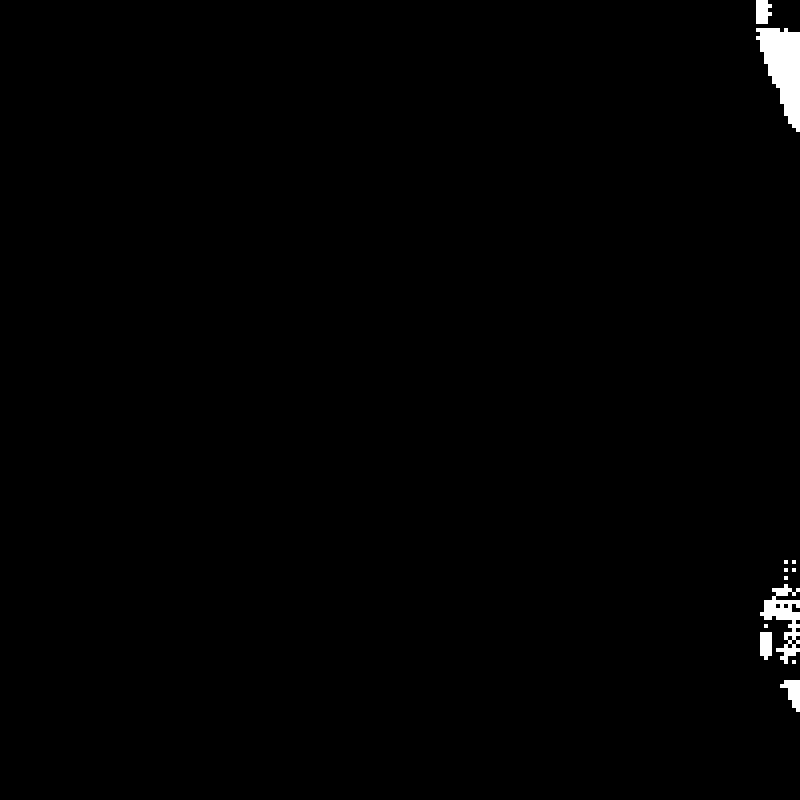}}
     \vspace{.1cm}
      \centerline{(2)}\medskip
\end{minipage}
\begin{minipage}[]{.155\linewidth}
    \centering
    \centerline{\includegraphics[width=2.8cm]{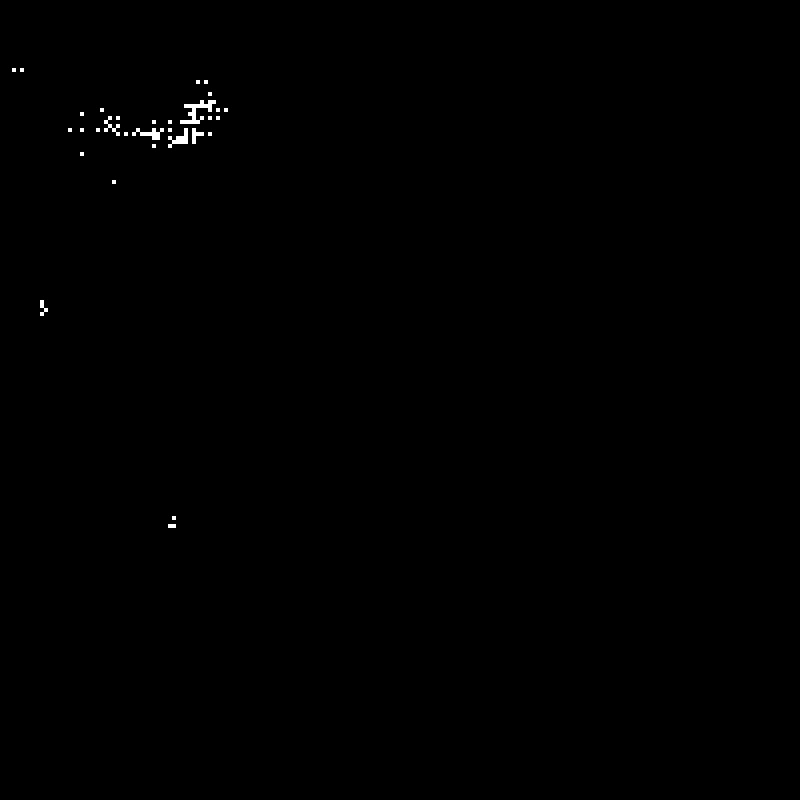}}
     \vspace{.1cm}
      \centerline{(3)}\medskip
\end{minipage}
\begin{minipage}[]{.155\linewidth}
    \centering
    \centerline{\includegraphics[width=2.8cm]{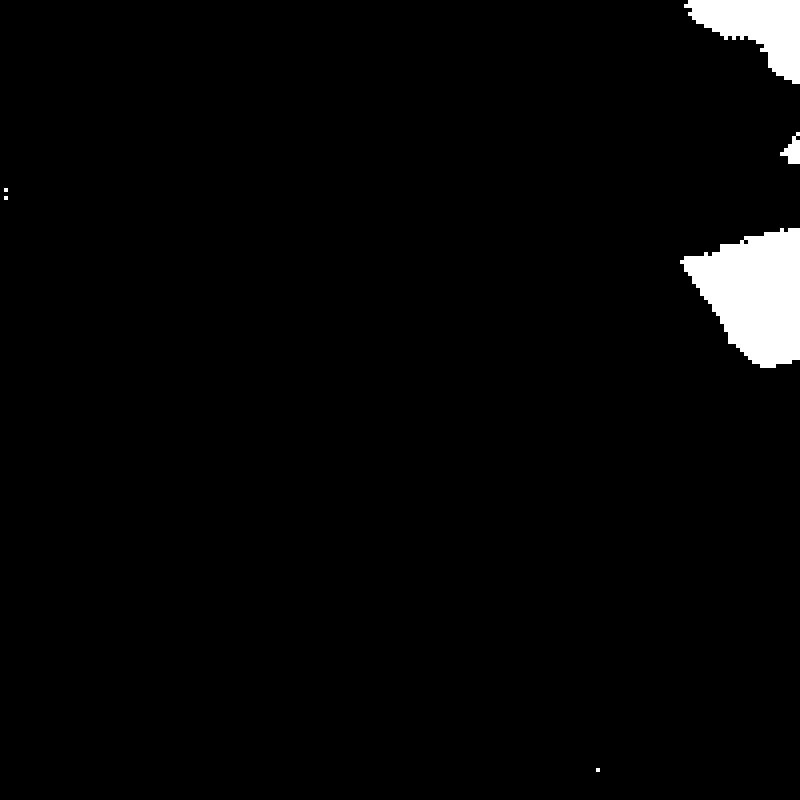}}
     \vspace{.1cm}
      \centerline{(4)}\medskip
\end{minipage}
\begin{minipage}[]{.155\linewidth}
    \centering
    \centerline{\includegraphics[width=2.8cm]{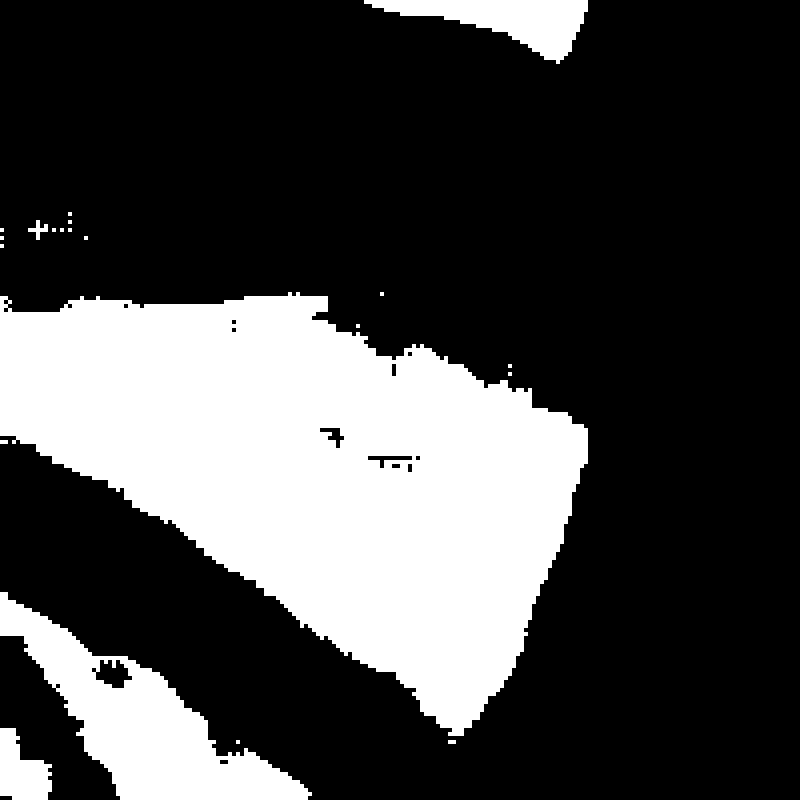}}
     \vspace{.1cm}
      \centerline{(5)}\medskip
\end{minipage}
\begin{minipage}[]{.155\linewidth}
    \centering
    \centerline{\includegraphics[width=2.8cm]{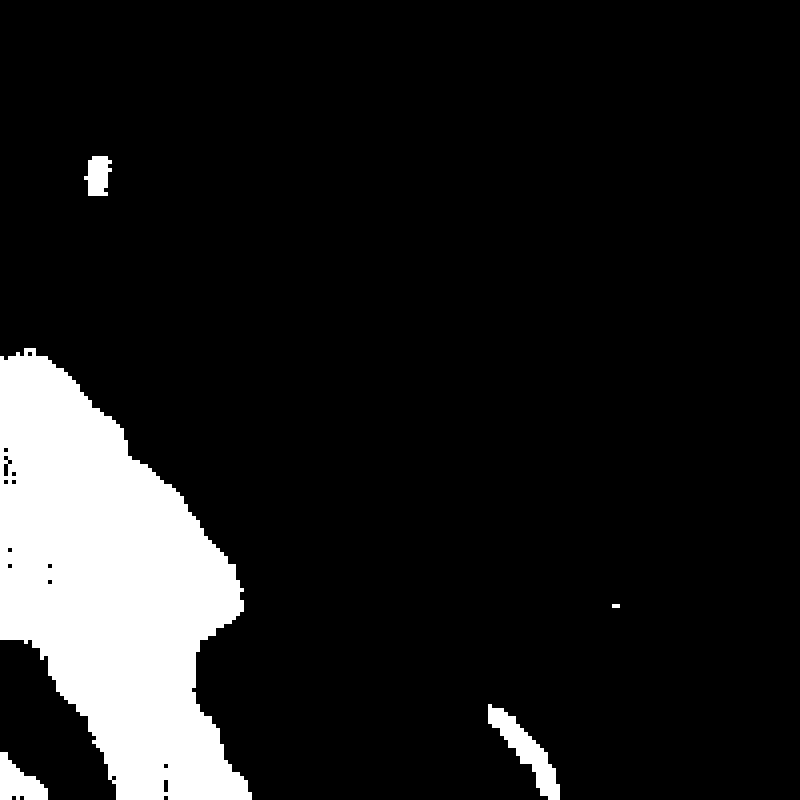}}
     \vspace{.1cm}
      \centerline{(6)}\medskip
\end{minipage}
\caption{\centering Patch similarity measurement. The reference patches are in the top row. Features are extracted from the images in the second row, and then each of them is compared to the one of the ``boat", the ``city", and the ``low city" reference patches. Only the features that have a cosine similarity of more than 0.8 with the reference patches are kept. }
\label{fig:pattern}
\end{figure*}

The experiment is conducted on the HRSID dataset \cite{wei2020hrsid}, which comprises multiple types of SAR sensors. In this case, we only use the Sentinel-1B satellite with HH polarization. Patches of size $64\times 64$ are extracted from the images with a stride of 4.  Patches with a cosine similarity measure higher than 0.8 relative to the reference patch are retained in the detection map, while others are discarded. Some results are displayed in Fig.~\ref{fig:pattern}. The three reference patches include a boat from the image (3) labeled "boat", a high-density urban area with roads and buildings from the image (4) labeled "city", and a low-intensity urban area from the image (5) labeled "urban".\\
\clearpage
\clearpage

When analyzing the similarity between the images and the "boat" patch, we observe that boats in images (1), (2), and (3) are detected. The patch size makes the detected area broader than the actual boat. Importantly, this method does not detect other bright scatterers from the city, which improves statistical detectors that rely solely on intensity. Traditional detectors might cluster boats and bright backscatterers like buildings into the same category due to their similar intensities. Our feature extractor allows the application of a standard detector on the features, providing richer information beyond pixel intensity alone.\\

When comparing similarity with the "city" patch, the area in image (4) from which the patch is extracted is accurately detected. Additionally, parts of the shore with bright backscatterers in images (4), (5), and (6) are also detected. However, the center bottom of the image (4) is not detected. It contains different types of backscatterers than in the "city" patch, and the road is not visible. This indicates that the network can differentiate between these structures.\\

For the "urban" reference patch, visually similar areas in images (4), (5), and (6) are well detected. However, some parts of the sea are wrongly detected in images (1) and (3). As observed in Section \ref{sec:seg} and in Table \ref{table:iou}, while the feature extractor facilitates the detection of different scatterers with similar intensity, it also complicates the detection of water from low-intensity urban areas or vegetation.

\section{Conclusion}\label{sec:conclusions}

In this paper, we have presented a novel SSL framework tailored specifically for SAR imagery, utilizing masked Siamese ViTs. We developed SAFE, a general SAR feature extractor capable of handling various SAR acquisition modes and resolutions, as well as both mono-channel SAR and PolSAR amplitude data. Our approach effectively addresses SAR imagery's unique challenges, such as speckle perturbation and multi-sensor resolution.

Our method was evaluated on multiple datasets and tasks, including target recognition with few-shot learning, segmentation, pattern detection, and visualization. These evaluations highlight the adaptability of our feature extractor to different modalities and conditions. Notably, SAFE has not been trained on the datasets used for segmentation and classification, yet it achieves performances that are on par with or better than state-of-the-art models trained specifically on those datasets.

The results indicate that SAFE can serve as a backbone for a wide range of SAR-based applications, enhancing their performance and reliability. Moreover, the adaptability and generalizability of our model underscore its potential for future applications. With the necessary resources, it would be possible to train SAFE on a larger scale, incorporating many more sensors and surface types.

\bibliographystyle{ieeetr}
\bibliography{refs}
\end{document}